\documentclass[12pt]{article}

\usepackage{amsmath,amssymb}
\usepackage{graphicx}
\usepackage[font={small,it}]{caption}
\usepackage[font={small,it}]{subcaption}
\usepackage{color}
\usepackage{dcolumn}
\usepackage{bm}
\usepackage{float}
\usepackage{hyperref} 
\usepackage{apacite}
\usepackage{makecell}
\usepackage{adjustbox}
\usepackage[linesnumbered,lined,ruled,vlined,noend]{algorithm2e}
\usepackage{hhline}
\usepackage{authblk}

\hypersetup{ colorlinks = true, citecolor = black, urlcolor = blue}
\newcommand{\qed}{\hfill \ensuremath{\Box}}

\definecolor{background-color}{gray}{0.98}
\newcommand{\br}[1]{\left\{#1\right\}}                            

\newcommand{\caras}{\textsc{Caratheodory}}\makeatletter

\newcommand{\Y}{\mathcal{X}}
\newcommand{\Q}{\mathbb{Q}}
\newcommand{\Z}{\mathbb{Z}}
\newcommand{\ff}{f}
\newcommand{\hh}{h}
\newcommand{\loss}{\mathrm{loss}}

\newcommand{\merge}{\textsc{merge}}
\newcommand{\reduce}{\textsc{reduce}}

\newcommand{\coresetAlg}{\textsc{CoresetAlg}}
\newcommand{\streamingAlg}{\textsc{Streaming-Coreset}}
\newcommand{\distributedAlg}{\textsc{DistributedAlg}}
\newcommand{\instream}{stream}

\newcommand{\pow}{\mathcal{P}}
\newcommand{\ones}{\textbf{1}}
\newcommand{\tovec}{\mathrm{vec}}
\newcommand{\dist}{\mathrm{dist}}

\newcommand{\REAL}{\ensuremath{\mathbb{R}}}
\providecommand{\norm}[1]{\left\lVert#1\right\rVert}

\renewcommand{\epsilon}{\varepsilon}
\newtheorem{theorem}{Theorem}

\newtheorem{definition}[theorem]{Definition}

\DeclareMathOperator*{\argmin}{arg\,min}
\DeclareMathOperator*{\argmax}{arg\,max}

\def\easy{\rlap{\protect\makebox{\color{red}$\diamond$}}}
\def\medium{\rlap{\protect\makebox{\color{red}$\diamond\diamond$}}}
\def\hard{\rlap{\protect\makebox{\color{red}$\diamond\diamond\diamond$}}}

\date{}
\begin{document}

\title{Introduction to Coresets: \\
Accurate Coresets}
\author{Ibrahim Jubran}
\author{Alaa Maalouf}
\author{Dan Feldman}
\affil{\texttt{\{ibrahim.jub, Alaamalouf12, dannyf.post\}@gmail.com}

The Robotics and Big Data Lab

Department of Computer Science

University of Haifa, Israel}
\maketitle


\begin{abstract}
A coreset (or core-set) of an input set is its small summation, such that solving a problem on the coreset as its input, provably yields the same result as solving the same problem on the original (full) set, for a given family of problems (models, classifiers, loss functions).  Over the past decade, coreset construction algorithms have been suggested for many fundamental problems in e.g. machine/deep learning, computer vision, graphics, databases, and theoretical computer science. This introductory paper was written following requests from (usually non-expert, but also colleagues) regarding the many inconsistent coreset definitions, lack of available source code, the required deep theoretical background from different fields, and the dense papers that make it hard for beginners to apply coresets and develop new ones.

The paper provides folklore, classic and simple results including step-by-step proofs and figures, for the simplest (accurate) coresets of very basic problems, such as: sum of vectors, minimum enclosing ball, SVD/ PCA and linear regression. Nevertheless, we did not find most of their constructions in the literature. Moreover, we expect that putting them together in a retrospective context would help the reader to grasp modern results that usually extend and generalize these fundamental observations. Experts might appreciate the unified notation and comparison table that links between existing results.

Open source code with example scripts are provided for all the presented algorithms, to demonstrate their practical usage, and to support the readers who are more familiar with programming than math.
\end{abstract}

\section{Introduction} \label{sec:intro}
Coreset (or core-set) is a modern data summarization that approximates the original data in some provable sense with respect to a (usually infinite) set of questions, queries or models and an objective loss/cost function. The goal is usually to compute the model that minimizes this objective function on the small coreset instead of the original (possibly big) data, without compromising the accuracy by more than a small multiplicative factor. Moreover, it has many other applications such as handling constraints, streaming , distributed data, parallel computation, model compression, parameter tuning, model selection and many more.

The simplest coreset is a (possibly weighted) \emph{subset} of the input data. The advantages of such subset coresets are: (i) preserved sparsity of the input, (ii) interpretability, (iii) coreset may be used (heuristically) for other problems, (iv) less numerical issues that occur when non-exact linear combination of points are used~\cite{maalouf2019fast}. Unfortunately, not all problems admit such a subset coreset, as we show throughout the paper.

Although coreset constructions are usually practical and not hard to implement, the theory behind them may be complicated and based on good understanding of linear algebra, statistics, probability, computational geometry and machine learning. Similarly to approximation algorithms in computer science, there are some generic techniques for coreset constructions, but many of their constructions are heavily tailored and related to the problem at hand and its existing solvers.
Furthermore, there are many inconsistent definitions of coresets in the papers. Nevertheless, it seems that after understanding the intuition and math behind simple coreset constructions, it is much easier to read modern academic papers and construct coresets for new problems.

To this end, this paper focuses only on what seems to be the simplest type of coresets, namely ``accurate coresets'', which do not introduce any approximation error when compressing the original data, but give accurate solutions.

Most of the coresets in this paper are easy to construct and may be considered as ``folklore" results. However, we did not find them in the literature, and we realized that many experts in the field are not familiar with them. Furthermore, since most of these results are easy to construct and explain, we found them to be suitable for tutorials, as the case in this paper. These results may also be of great interest to people from various fields of study, who may not be familiar even with the simple techniques presented in this paper. We assume no previous knowledge except from basic linear algebra, and therefore we target both experts and beginners in the field, as well as data scientists and analysts. To better understand the results presented in this paper and to encourage people to use them, we provide full open-source code for these results~\cite{opencode}.

Another motivation of this introductory survey is to show the many possible different definitions of coresets and the resulting different constructions, as well as summarizing them in a single place.


Table~\ref{table:ourContrib} summarises the different accurate coresets that we present in this paper.

\begin{table}[ht]
\begin{adjustbox}{width=1\textwidth}
\small
\begin{tabular}{ | c | c | c | c | c | c | c | c | c | c |}
\hline
\textbf{Name} & \textbf{\makecell{Input Weighted Set\\$(P,w)$ of size $|P|=n$}} & \textbf{\makecell{Query Set $\Y$}} & \textbf{cost function $\ff:P\times\Y$} & \textbf{\makecell{$\loss$ for\\$f(p,x)$\\over $p\in P$}} & \textbf{\makecell{Coreset $C$}} & \textbf{\makecell{Coreset Weights}} & \textbf{\makecell{Const.\\time}} & \textbf{\makecell{Query\\time}} & \textbf{Section}\\
\hline
\makecell{$1$-Center} & \makecell{$P\subseteq \ell \subseteq \REAL^d$\\$w\equiv 1$} & $\Y = \REAL^d$ & $\ff(p,x) = \norm{p-x}$ & $\norm{\cdot}_\infty$ & $\makecell{C\subseteq P\\ |C|=2}$ & $u\equiv 1$ & $O(n)$ & $O(d)$ & \ref{1center}\\
\hline
\makecell{Monotonic\\function} & \makecell{$P\subseteq \REAL$\\$w\equiv 1$} & \makecell{$\Y$ = \{$g \mid g$ is monotonic\\decreasing/increasing or\\increasing and then\\decreasing function\}} & $\ff(p,g) = g(p)$ & $\norm{\cdot}_\infty$ & $\makecell{C\subseteq P\\ |C|=2}$ & $u\equiv 1$ & $O(n)$ & $O(1)$ & \ref{monotonicFunc}\\
\hline
\makecell{Vectors\\sum (1)} & \makecell{$P\subseteq \REAL^d$\\$w:P\to\REAL$} & $\Y = \REAL^d$ & $\ff(p,x) = p-x$ & $\Sigma$ & $\makecell{C\subseteq \REAL^d\\ |C|=1}$ & $u\equiv \sum_{p\in P}w(p)$ & $O(n)$ & $O(d)$ & \ref{shiftedMean1}\\
\hline
\makecell{Vectors\\sum (2)} & \makecell{$P\subseteq \REAL^d$\\$w:P\to\REAL$} & $\Y = \REAL^d$ & $\ff(p,x) = p-x$ & $\Sigma$ & $\makecell{C\subseteq P\\ |C| \leq d+1}$ & \makecell{$u:C\to\REAL$\\$\displaystyle\sum_{p\in C}u(p)=\sum_{p\in P}w(p)$} & $O(nd^2)$ & $O(d^2)$ & \ref{shiftedMean2}\\
\hline
\makecell{Vectors\\sum (3)} & \makecell{$P\subseteq \REAL^d$\\$w:P\to [0,\infty)$} & $\Y = \REAL^d$ & $\ff(p,x) = p-x$ & $\Sigma$ & $\makecell{C\subseteq P\\ |C|\leq d+2}$ & \makecell{$u:C\to \left[0,\displaystyle\sum_{p\in P}w(p)\right]$\\$\displaystyle\sum_{p\in C}u(p)=\sum_{p\in P}w(p)$} & \makecell{$O\big(\min \{n^2d^2$,\\$nd +d^4log{n}\}\big)$} & $O(d^2)$ & \ref{shiftedMean3}\\
\hline
\makecell{$1$-Mean (1)} & \makecell{$P\subseteq \REAL^d$\\$w:P\to \REAL$} & $\Y = \REAL^d$ & $\ff(p,x) = w(p)\norm{p-x}^2$ & $\norm{\cdot}_1$ & \makecell{$C\subseteq\REAL^d\times \Z\times \REAL$\\$|C|=3$\\Different loss} & $unweighted$ & $O(nd)$ & $O(d)$ & \ref{1mean1}\\
\hline
\makecell{$1$-Mean (2)} & \makecell{$P\subseteq \REAL^d$\\$w:P\to\REAL$} & $\Y = \REAL^d$ & $\ff(p,x) = w(p)\norm{p-x}^2$ & $\norm{\cdot}_1$ & \makecell{$C\subseteq P$\\$|C| \leq d+2$} & \makecell{$u:C\to \REAL$\\$\displaystyle\sum_{p\in C}u(p)=\sum_{p\in P}w(p)$\\$\displaystyle\sum_{p\in C}u(p)\norm{p}^2=\sum_{p\in P}w(p)\norm{p}^2$} & $O(nd^2)$ & $O(d^2)$ & \ref{1mean2}\\
\hline
\makecell{$1$-Mean (3)} & \makecell{$P\subseteq \REAL^d$\\$w:P\to [0,\infty)$} & $\Y = \REAL^d$ & $\ff(p,x) = w(p)\norm{p-x}^2$ & $\norm{\cdot}_1$ & \makecell{$C\subseteq P$\\$|C|\leq d+3$} & \makecell{$u:C\to \left[0,\displaystyle\sum_{p\in P}w(p)\right]$\\$\displaystyle\sum_{p\in C}u(p)=\sum_{p\in P}w(p)$\\$\displaystyle\sum_{p\in C}u(p)\norm{p}^2=\sum_{p\in P}w(p)\norm{p}^2$} & \makecell{$O\big(\min \{n^2d^2$,\\$nd +d^4log{n}\}\big)$} & $O(d^2)$ & \ref{1mean3}\\
\hline
\makecell{$1$-Segment} & \makecell{$P = \br{(t_i\mid p_i)}_{i=1}^n \subseteq \REAL^{d+1}$\\$w:P \to [0,\infty)$} & $\Y = \br{g\mid g:\REAL\to\REAL^d}$ & $\ff((t,p),g) = \norm{p-g(t)}^2$ & $\norm{\cdot}_1$ & \makecell{$C \subseteq \REAL^{d+1}$\\$|C|=d+2$} & \makecell{$u \equiv 1$} & $O(nd^2)$ & $O(d^2)$ & \ref{1segment}\\
\hline
\makecell{Matrix\\$2$-norm (1)} & \makecell{$P \subseteq \REAL^{d}$\\$w:P\to [0,\infty)$} & $\Y = \REAL^d$ & $\ff(p,x) = (p^Tx)^2$ & $\norm{\cdot}_1$ & \makecell{$C\subseteq \REAL^d$\\$|C|=d$} & $u\equiv 1$ & $O(nd^2)$ & $O(d^2)$ & \ref{matrix2norm1}\\
\hline
\makecell{Matrix\\$2$-norm (2)} & \makecell{$P \subseteq \REAL^{d}$\\$w:P\to [0,\infty)$} & $\Y = \REAL^d$ & $\ff(p,x) = (p^Tx)^2$ & $\norm{\cdot}_1$ & \makecell{$C\subseteq P$\\$|C|\leq d^2+1$} & \makecell{$u:C\to\left[0,\displaystyle\sum_{p\in P}w(p)\right]$\\$\displaystyle\sum_{p\in C}u(p) = \sum_{p\in P}w(p)$} & \makecell{$O\big(\min \{n^2d^4$,\\$nd^2 +d^8log{n}\}\big)$} & $O(d^3)$ & \ref{matrix2norm2}\\
\hline
\makecell{Least\\Mean\\Squares} & \makecell{$P = \br{(a_i^T\mid b_i)}_{i=1}^n \subseteq \REAL^{d+1}$\\$w:P\to [0,\infty)$} & $\Y = \REAL^d$ & $\ff((a^T\mid b),x) = (a^Tx-b)^2$ & $\norm{\cdot}_1$ & \makecell{$C\subseteq P$\\$|C|\leq (d+1)^2+1$} & \makecell{$u:C\to\left[0,\displaystyle\sum_{p\in P}w(p)\right]$\\$\displaystyle\sum_{p\in C}u(p) = \sum_{p\in P}w(p)$} & \makecell{$O\big(\min \{n^2d^4$,\\$nd^2 +d^8log{n}\}\big)$} & $O(d^3)$ & \ref{sec:LMS}\\
\hline
\end{tabular}
\end{adjustbox}
\caption{\textbf{Coresets that are presented in this paper.} The input set, the query set, and the roles of the functions $\ff$ and $\loss$ are as defined in Definition~\ref{def::query space}. The first and second arguments of the function $f$ are elements from the input set and the query set respectively. We assume that the input set is of size $|P|=n$, and we wish to compute the loss over the $n$ fitting errors that are defined by $f$, each input point and a given query.}
\label{table:ourContrib}
\end{table}

\section{Preliminalies} \label{sec:preLim}
In this section, we give basic notations and definitions that will be used throughout this paper.

The set of all real numbers is denoted by $\REAL$. We denote $[n]=\br{1,\cdots,n}$ for every integer $n \geq 1$ by $\norm{p}=\norm{p}_2=\sqrt{p_1^2+\ldots+p_d^2}$ the $\ell_2$ norm of a point $p=(p_1,\cdots,p_d) \in \REAL^d$, by $\norm{p}_q = \sqrt[1/q]{\sum_{i=1}^d |p_i|^q}$ the $\ell_q$ norm of $p$ for every $q>0$, by $\norm{p}_{\infty} = \max_{i}|p_i|$ the $\ell_\infty$ norm, and by $\norm{A}_F = \sqrt{\sum_{i=1}^m\sum_{j=1}^n a_{ij}^2}$ the Frobenius norm of a matrix $A \in \REAL^{m\times n}$, where $a_{ij}$ is the $j$th entry at the $i$th row of $A$. The $d$-dimensional identity matrix is denoted by $I_d \in \REAL^{d\times d}$. For a function $\ff$ we denote $\ff^2(\cdot,\cdot)=(\ff(\cdot,\cdot))^2$. For a set $Z$ of elements, we denote by $\pow(Z)$ the power set of $Z$, i.e., the set of all subsets of $Z$.


A \textit{weighted set} is a pair $P'=(P,w)$ where $P$ is a set of items called \emph{points}, and $w:P\to \REAL$ is a function that maps every $p\in P$ to $w(p) \in \REAL$, called the \emph{weight} of $p$. A \emph{weighted point} is a weighted set of size $|P|=1$.
A weighted set $(P,\mathbf{1})$ where $\mathbf{1}$ is the weight function $w:P\to\br{1}$ that assigns $w(p)=1$ for every $p\in P$ may be denoted by $P$ for short.

In order to have a unified framework, we make the following definition of a query space, which will simplify and unify the definitions of every example coreset presented in this paper. A query space basically includes all the ingredients needed to define a coreset for a new problem that we wish to tackle.

\begin{definition}[query space]\label{def::query space}
Let $\Y$ be a (possibly infinite) set called \emph{query set},  $P'=(P,w)$ be a weighted set called the \emph{input set}, $\ff:P\times \Y\to [0,\infty)$ be called a \emph{cost function}, and $\loss$ be a function that assigns a non-negative real number for every real vector.
The tuple $(P,w,\Y,\ff,\loss)$ is called a \emph{query space}. For every weighted set $C'=(C,u)$ such that $C=\br{c_1,\cdots,c_m}$,
and every $x\in \Y$ we define the overall fitting error of $C'$ to $x$ by
\[
\ff_{\loss}(C',x):=\loss((w(c)\ff(c,x))_{c\in C})=\loss(w(c_1)\ff(c_1,x),\cdots,w(c_m)\ff(c_m,x)).
\]
\end{definition}

An accurate coreset that approximates a set of models (queries) for a specific problem is defined as follows:

\begin{definition}[accurate coreset\label{coresetdef3}]
Let $P' = (P,w)$ be a weighted set and $(P,w,\Y,\ff,\loss)$ be a query space. The weighted set $C'$ is called an \emph{accurate coreset} for $(P,w,\Y,\ff,\loss)$ if for every $x\in \Y$ we have
\[
\ff_{\loss}(P',x) = \ff_{\loss}(C',x).
\]
\end{definition}

\section{Accurate Coresets} \label{sec:AccurateCoreset}
In what follows, each subsection presents an accurate coreset construction for a given query space. Each section is marked with one of the following difficulty indicators: \easy\quad (easy), \medium\quad (intermediate) or \hard\quad\quad (advanced).

\subsection{$1$-Center \easy}\label{1center}
Suppose that we want to open a shop on our street that will be close to all the residents in the street, and suppose that the street is represented by a linear segment, say the $x$-axis, while the residents are represented by points on this segment. Since we want to be close to all the potential $n\geq1$ residents in the street, if we decide to open the shop at some location, our loss will be measured as the distance to the farthest resident. Suppose that tomorrow we will be given few locations to choose from to position our store. Can we pre-process the given positions of the residents so that computing the cost (farthest resident) from the suggested store location will take only $O(1)$ time for each suggested location? That is, constant time that is independent of the number of residents $n$?

\begin{figure}[h]
	\begin{subfigure}[t]{0.5\textwidth}
    \includegraphics[width=\textwidth]{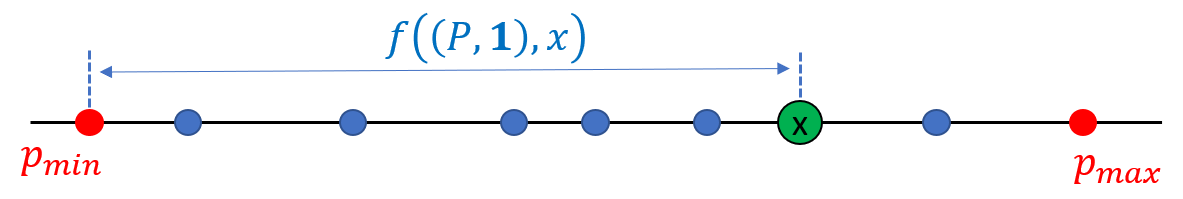}
    \caption{An input  set $P\subseteq \REAL$ (in blue) on the $x$-axis and a query point $x\in \REAL$ (in green).}
    \label{fig:1center1D}
    \end{subfigure}
    ~
    \begin{subfigure}[t]{0.5\textwidth}
    \includegraphics[width=\textwidth]{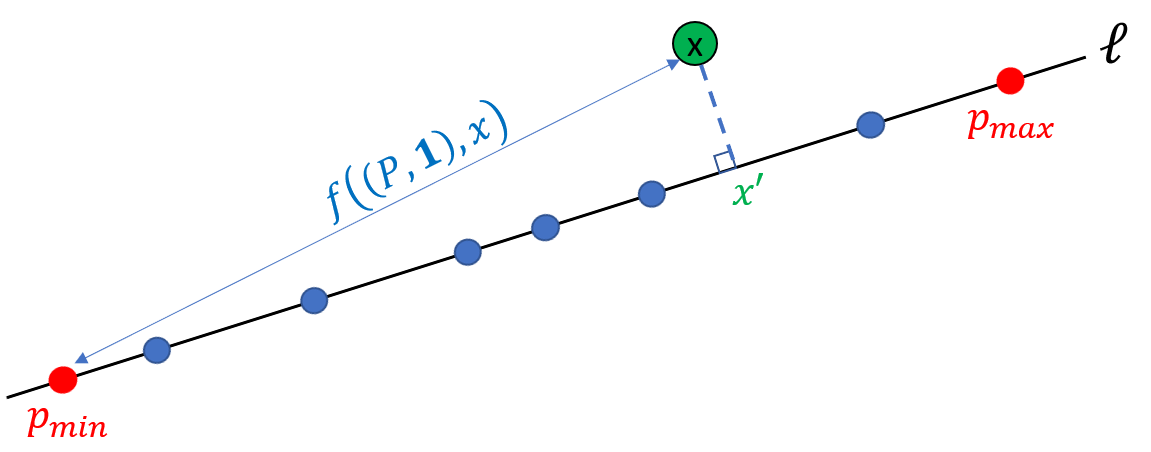}
    \caption{A line $\ell$ in $\REAL^d$, an input set $P\subseteq \ell$ (in blue) and a query point $x\in \REAL^d$ (in green).}
    \label{fig:1centerRd}
    \end{subfigure}
    \caption{\textbf{An accurate coreset for the $1$-center query space (red points).} Both in Fig.~\ref{fig:1center1D} and Fig.~\ref{fig:1centerRd}, the farthest point $p\in P$ from the query $x$ is either the first point ($p_{min}$) or the last point ($p_{max}$) on the line, i.e., $\ff_{\loss}((P,\ones),x) = \max\br{\norm{p_{min}-x},\norm{p_{max}-x}}$.}
    \label{fig:1center}
\end{figure}

Formally, in the query-space notation from Definition~\ref{def::query space}, we have that
\[
P = \br{p_1,\cdots,p_n}\subseteq \REAL, w\equiv 1, \Y = \REAL, \ff(p,x) = |p-x|, \loss(\cdot) = \norm{\cdot}_\infty.
\]
Our goal is to compute a data structure (accurate coreset) $C$ such that for every number $x\in \Y$ we can compute
\[
\ff_{\loss}((P,\ones),x)= \big|\big|\big(|p_1-x|,\cdots,|p_n-x|\big)\big|\big|_\infty =\max_{p\in P} |p-x|
\]
in $O(1)$ time using only $C$. This can be easily done by observing that for every $x\in\Y$, the farthest point from $x$ is either the smallest point $p_{min}$ or the largest point $p_{max}$ in $P$; See Fig.~\ref{fig:1center1D}. That is, simply choosing $C=\br{p_{min},p_{max}} \subseteq P$ yields
\[
\ff_{\loss}((P,\ones),x)= \max_{p\in P} |p-x| = \max \br{|p_{min}-x|, |p_{max}-x|} = \ff_{\loss}((C,\ones),x),
\]
i.e., the distance to the farthest input point from $x$ is either the distance to the leftmost or the rightmost point of $P$.

Even for the case that $\Y = \REAL^d$ and where $P$ is contained on a line $\ell$ in $\REAL^d$, we would still have $\ff_{\loss}((P,\ones),x)=\ff_{\loss}((C,\ones),x)$ where $C$ contains the two edge points on $\ell$; See Fig.~\ref{fig:1centerRd}. Indeed, denote by $x'$ the projection (closest point) of $x$ onto $\ell$ to obtain by the Pythagorean Theorem that
\[
\begin{split}
\ff^2_{\loss}((P,\ones),x)&=\max_{p\in P}\norm{p-x}^2
=\max_{p\in P}\norm{p-x'}^2+\norm{x'-x}^2\\
&=\max_{p\in C}\norm{p-x'}^2+\norm{x'-x}^2
=\max_{p\in C}\norm{p-x}^2
=\ff^2_{\loss}((C,\ones),x).
\end{split}
\]
See implementation of function \texttt{one\_center} in~\cite{opencode}.

The solution in this section does not generalize to an arbitrary set of points in $\REAL^d$. In fact, the following set of points on the plane does not have any subset which is an accurate coreset.
Let $P\subseteq\REAL^2$ denote $n$ points on the unit circle in the plane, and let $C\subset P$. For $p\in P\setminus C$ and $x = -p$ we have that $\ff_\loss((P,\ones),x) = \ff_\loss(\br{p},x) \neq \ff_\loss((C,\ones),x)$.
Hence, there is no subset $C$ which is a coreset for $P$ in this sense.
Nevertheless approximated coresets can be found in~\cite{paul2014visual} and references therein.

Furthermore, the above solution does not generalize for the case where the input is weighted, even for $d=1$, i.e., there is a weighted set of points $(P,w)$ where $P\subseteq \REAL$ such that for every $p\in P$ there is $x\in \REAL$ that satisfies $\ff_\loss((P,w),x) = w(p)|p-x|$. In other words, assume $p_j \in P$ was not chosen for the coreset $(C,u)$, then there is some query $x \in \REAL$ such that $\ff_\loss((P,w),x) = w(p_j)|p_j-x| \neq \ff_\loss((C,u),x)$.

\begin{figure}
  \centering
  \includegraphics[width=0.5\textwidth]{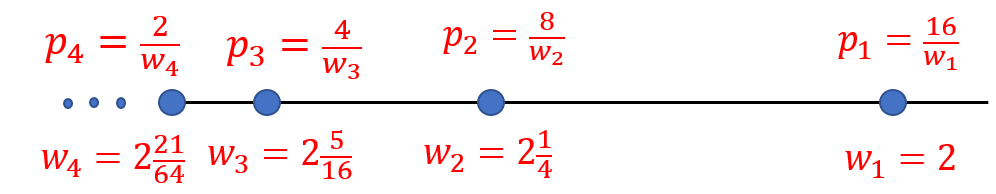}
  \caption{\textbf{No accurate coreset exists for the weighted 1-center problem.} A weighted set $(P,w)$, $P \subseteq \REAL$, where $w(p_1) = 2$, $w(p_i) = w_{p_{i-1}} + \left(\frac{1}{4}\right)^{i-1}$, and $p_i = \frac{2^{5-i}}{w(p_i)}$. For every $p_i \in P$, we can compute a query $x_i \in \REAL$ where $x_i = -2^{4+i}+1$ such that $\ff_\loss((P,w),x_i) = \max_{p\in P} w(p)|p-x_i| = w(p_i)|p_i-x_i|$.}
  \label{fig:weighted1Center}
\end{figure}

We now construct such an example. Let $(P,w)$ be a weighted set of $|P| = n$ points where
\[
w(p_1) = 2, \quad w(p_i) = w_{p_{i-1}} + \left(\frac{1}{4}\right)^{i-1},
\]
and
\[
p_i = \frac{2^{5-i}}{w(p_i)},
\]
as illustrated in Fig.~\ref{fig:weighted1Center}.
For every $p_i \in P$, by defining $x_i = -2^{4+i}+1$, it is easy to verify that $\ff_\loss((P,w),x_i) = \max_{p\in P}w(p)|p-x_i| = w(p_i)|p_i-x_i|$. Therefore, any coreset for this problem must include all the input points in $(P,w)$. Thus there is no accurate coreset for the weighted $1$-center problem, even in $1$-dimensional space.

\subsection{Monotonic functions \easy} \label{monotonicFunc}
\begin{figure}[h]
\centering
    \includegraphics[width=0.7\textwidth]{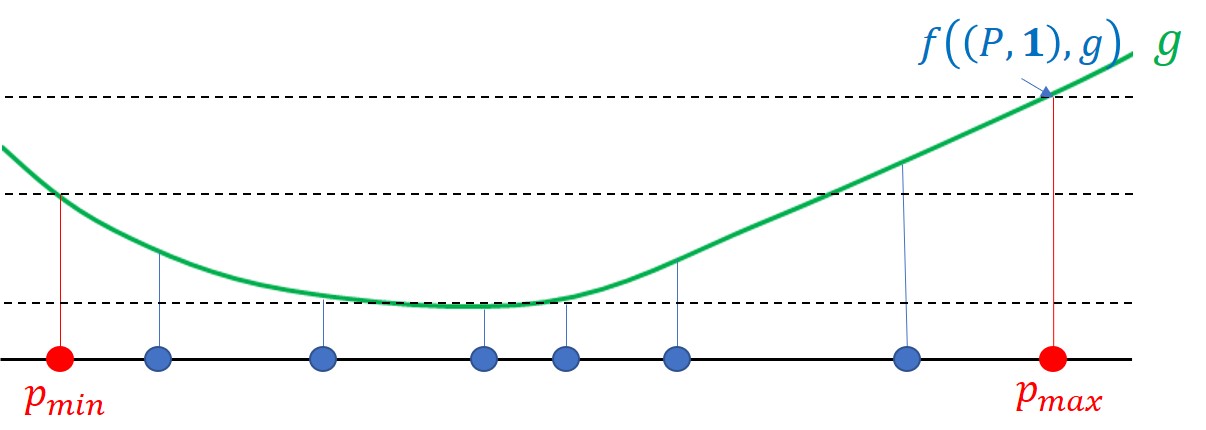}
    \caption{\textbf{Accurate coreset for monotonic functions.} A set $P\subseteq \REAL$ and a ``decreasing then increasing'' monotonic function $g:\REAL\to [0,\infty)$. The point that maximizes $g(p)$ over every $p\in P$ is either $p_{min}$ or $p_{max}$, i.e., $\ff_{\loss}((P,\ones),g) = \max\br{g(p_{min}),g(p_{max}})$.}
    \label{fig:MonotonicFunction}
\end{figure}

What if the function $\ff(p,x) = \norm{p-x}$ from Section~\ref{1center} is not an Euclidean distance, but a function of this distance $g(\norm{p-x})$, where $g(y)=y^2$ so $\ff(p,x)=\norm{p-x}^2$ is the squared Euclidean distance from $x$, or $g(y)=\min\br{y,1}$ so $\ff(p,x)=\min\br{1,\norm{p-x}^2}$? The latter one is called an M-estimator and is robust to points that are very far from $x$ (outliers). It turns out that the coreset from the previous section holds for the following cases.
Consider the query space $(P,w,\REAL^d,\ff, \loss)$ where $P = \br{p_1,\cdots,p_n}\subseteq\REAL$, $w \equiv 1$, the query set $\Y$ is the union over every function $g:\REAL\to[0,\infty)$ that is a non-negative decreasing, increasing, or decreasing and then increasing monotonic function, $\ff(p,g) = g(p)$, and $\loss(\cdot) = \norm{\cdot}_\infty$. Note that here every query is actually a function and not a point.
Hence,
\[
\ff_{\loss}((P,\ones),g)=\norm{(g(p_1),\cdots,g(p_n))}_\infty=\max_{p\in P}g(p).
\]
Again, the main observation is that the maximum value of $g(p)$ over $p\in P$ is obtained in one of the points $p_{max} \in \displaystyle\argmax_{p\in P}p$ or $p_{min} \in \displaystyle\argmin_{p\in P}p$; See Fig.~\ref{fig:MonotonicFunction}. Therefore, the coreset $C = \br{p_{min},p_{max}}$ from Section~\ref{1center} is also valid here.

\subsection{Vectors sum \medium} \label{shiftedMean1}
The accurate coreset for vectors sum example presented in Sections~\ref{shiftedMean1}--\ref{shiftedMean3} is a warm-up example, that will be used in later sections.

Consider the query space $(P,w,\REAL^d,\ff, \loss)$ where $P = \br{p_1,\cdots,p_n}$ is a set of $n$ points in $\REAL^d$, $w:P\to \REAL$, $\Y = \REAL^d$, $\ff(p,x) = p-x$ and $\loss$ maps every tuple $(v_1,v_2,\cdots,)$ of vectors to their sum $\sum_i v_i$ (which is also a vector). In this section, unlike other sections, the function $\ff$, as well as $\loss$, return a vector and not a positive scalar as in Definition~\ref{def::query space}. This query set defines the weighted mean or sum of the differences $(p-x)$ over $p\in P$, i.e.,
\[
\ff_{\loss}((P,w),x) = \sum_{p\in P}w(p)(p-x).
\]

It is easy to see that there is a coreset of a single weighted point to this query space. Indeed, let
\[
c=\frac{\sum_{p\in P}w(p)p}{\sum_{p\in P}w(p)},
\]
$C=\br{c}$ and $u(c) = \sum_{p\in P}w(p)$. Here we assume that $\sum_{p\in P}w(p)\neq 0$. Otherwise, set $c=\sum_{p\in P}w(p)p$ and $u(c)=1$.

We now have that the weighted mean of $(c-x)$ over $c\in C$ (a single point) is
\[
\begin{split}
\ff_{\loss}((C,u),x) & = u(c)(c-x) = \sum_{p\in P}w(p)(c-x)=\sum_{p\in P}w(p)p-x\cdot\sum_{p\in P}w(p)\\
& = \sum_{p\in P}w(p)p-x\sum_{p\in P}w(p) = \sum_{p\in P}w(p)(p-x) =\ff_{\loss}((P,w),x).
\end{split}
\]

See implementation of function \texttt{vectors\_sum\_1} in~\cite{opencode}.
\subsubsection{Subset coreset} \label{shiftedMean2}
The coreset for vectors sum in the previous section was not a subset of the input set $P$, i.e., $C = \br{c} \not\subseteq P$. In this section, the goal is to compute a weighted set $C' = (C,u)$ where $C \subseteq P$ is a subset of the input such that $\ff_{\loss}(C',x) = \ff_{\loss}(P',x)$. The motivation of a subset coreset is explained in Section~\ref{sec:intro}.
Let $\hat{p} = (p^T \mid 1)^T \in \REAL^{d+1}$ for every $p\in P$ denote the concatenation of $p$ with $1$, and let $\hat{P} = \br{\hat{p} \mid p \in P}$.

Given a distinct pair of points $p,q$ on a line in $\REAL^d$, we can describe any third point $z$ on the line as a linear combination of $p$ and $q$. More generally, every point in $\REAL^d$ can be described by $d$ independent vectors (points). Recall that in linear algebra such a set of $d$ vectors is called \emph{a basis} of $\REAL^d$.
Specifically, since $\sum_{\hat{p}\in \hat{P}} w(p)\hat{p} \in \REAL^{d+1}$ is spanned by (i.e., a linear combination of) the points of $\hat{P}$, there exists a set $\hat{C} \subseteq \hat{P}$ of $|\hat{C}| = d+1$ points, and a weight function $\hat{u}:\hat{C} \to \REAL$ such that
\begin{equation} \label{eqC21}
\sum_{\hat{p} \in \hat{C}} \hat{u}(\hat{p})\hat{p} = \sum_{\hat{p}\in \hat{P}} w(p)\hat{p} = \left(\sum_{p\in P}w(p)p \Bigm\vert \sum_{p\in P}w(p)\right).
\end{equation}
The set $\hat{C}$ can be computed as explained in Section~\ref{sec:LA}.
Here we assume $|P|>d+1$, otherwise we let $(C,u) = (P,w)$ be our coreset, since $P$ is already small. More generally, it is not hard to verify that $|C|$ can be small as the rank of the matrix whose rows are the points in $\hat{P}$.

Let $C = \br{p \in P \mid \hat{p} \in \hat{C}}$ and $u:C\to \REAL$ such that $u(p) = \hat{u}(\hat{p})$ for every $p \in C$.
It now follows that
\begin{equation} \label{eqC22}
\sum_{p \in C} u(p)(p \mid 1) = \sum_{\hat{p} \in \hat{C}} \hat{u}(\hat{p})\hat{p} = \left(\sum_{p\in P}w(p)p \Bigm\vert \sum_{p\in P}w(p)\right),
\end{equation}
where the first equality is by the definition of $u$ and $C$, and the second equality is by~\eqref{eqC21}.
By~\eqref{eqC22} we obtain that
\begin{equation}\label{eqC23}
\sum_{p \in C} u(p)p = \sum_{p\in P}w(p)p \text{\quad and \quad} \sum_{p \in C} u(p) = \sum_{p\in P}w(p).
\end{equation}

Hence, the weighted mean of $(p-x)$ over $p\in C$ is
\[
\begin{split}
\ff_{\loss}((C,u),x)&=\sum_{p \in C}u(p)(p-x)= \sum_{p \in C}u(p)p-x\sum_{p \in C}u(p)\\&= \sum_{p\in P}w(p)p-x\cdot\sum_{p\in P}w(p)= \sum_{p\in P}w(p)(p-x)=\ff_{\loss}(P',x),
\end{split}
\]
where the third equality is by~\eqref{eqC23}.

Observe that by~\eqref{eqC23} we obtain that the sum of weights of the coreset is the same as the sum of original weights. We thus have that $C' = (C,u)$ is an accurate coreset of size $|C| \leq d+1$, which is also a subset of the input set $P$, for the given query space.

\subsubsection{Subset Coreset of bounded positive weights} \label{shiftedMean3}
In the coreset of the previous section, the coreset's weights might be both negative and unbounded, even if the weights of the input points are bounded. This may cause serious issues as explained in~\cite{maalouf2019fast}.
In this section we prove that if the input weights are non-negative and sum to one, i.e., $w:P\to [0,1]$ and $\sum_{p\in P} w(p) = 1$, then there is an accurate coreset $C' = (C,u)$, where $C \subseteq P$ consists of $|C| = d+2$ points which is larger by one than the previous coreset, but has a non-negative weight function $u:C\to[0, 1]$  which also sums to one, i.e., $\sum_{p\in C} u(p) = 1$, instead of the previous unbounded weight function. This means that the weights are both non-negative and cannot be arbitrarily large, which reduces numerical issues. We then show how to naturally extend this result for the case where the input weights are non-negative but do not necessarily sum to one. In this generalized case, we compute an accurate coreset $(C,u)$ also of size $|C|=d+1$ such that $u:C\to [0,\sum_{p\in P} w(p)]$ and $\sum_{p\in P} w(p) = \sum_{p\in C} u(p)$.

\begin{figure}[h]
\centering
    \includegraphics[width=0.5\textwidth]{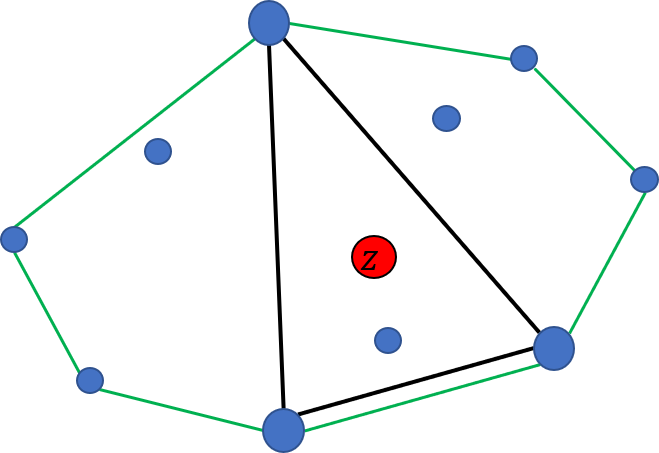}
    \caption{\textbf{Caratheodory's Theorem.} A set $P\subseteq \REAL^2$ (blue points), its mean $z$ (red point), and the convex hull of $P$ (green segments). The mean $z$ is contained inside the convex hull of $P$. There exists a set $\hat{P}\subseteq P$ of $|\hat{P}| = d+1 = 3$ points (bigger blue points) such that $z$ lies inside the convex hull of $\hat{P}$ (the black lines).}
    \label{fig:convexHull}
\end{figure}

\paragraph{Input weights are non-negative and sum to one.}
Let $\hat{p} = (p \mid 1)^T \in\REAL^{d+1}$ for every $p\in P$, and let $\hat{P} = \br{\hat{p} \mid p \in P}$ as in the previous example.
Observe that if $P$ is a set of points on a line, its mean $z$ must lie in the interval between the rightmost and leftmost points $p$ and $q$, respectively, of $P$. This implies that the mean is a convex combination of $p$ and $q$, i.e., $z=w_1p+w_2q$ for some $w_1,w_2\geq 0$ such that $w_1+w_2=1$ and $w_1,w_2\geq 0$.
For a set $P$ of points on the plane, the mean of $P$ is inside the convex hull of $P$, which is the smallest polygon that contains $P$, and there must be a triangle whose vertices are in $P$, that also contains $z$; See Fig.~\ref{fig:convexHull}.

More generally, Caratheodory's Theorem~\cite{caratheodory1907variabilitatsbereich,cook1972caratheodory} states that if a point $z$ lies inside a convex hull of the set $\hat{P} \subseteq \REAL^{d'}$, then $z$ also lies inside the convex hull (i.e., it is a convex combination) of at most $d'+1$ points in $\hat{P}$. Hence, $z$ can be expressed as a convex combination of at most $d'+1$ points in $\hat{P}$.
When the input weights sum to one, i.e., $\sum_{p\in P}w(p) = 1$, the weighted mean $\sum_{\hat{p}\in \hat{P}}w(p)\hat{p}$ of $\hat{P}$ lies inside the convex hull of $\hat{P}$.
Therefore, since $\hat{P} \subseteq \REAL^{d'} = \REAL^{d+1}$ and each point $\hat{p}$ is given a weight of $w(p)$ where $\sum_{\hat{p} \in \hat{P}}w(p) = 1$, by Caratheodory's theorem there is a subset $\hat{C}\subseteq \hat{P}$, $|\hat{C}| = d'+1 = d+2$ and $\hat{u}:\hat{C}\to [0,1]$ such that
\begin{equation}\label{eqC31}
\sum_{p\in \hat{C}}\hat{u}(p)p=\sum_{\hat{p}\in \hat{P}}w(p)\hat{p}= \left(\sum_{p\in P}w(p)p \Bigm\vert \sum_{p\in P}w(p)\right),
\end{equation}
where the second equality is by the definition of $\hat{p}$. $\hat{C}$ and $\hat{u}$ can be computed as explained in Section~\ref{sec:cara}.

Let $C = \br{p \in P \mid \hat{p} \in \hat{C}}$ and $u:C \to [0,1]$ such that $u(p) = \hat{u}(\hat{p})$ for every $p\in C$.
It now follows that
\begin{equation}\label{eqC32}
\sum_{p \in C} u(p)(p \mid 1) = \sum_{p \in \hat{C}} \hat{u}(\hat{p})\hat{p} = \left(\sum_{p\in P}w(p)p \Bigm\vert \sum_{p\in P}w(p)\right),
\end{equation}
where the first equality holds by the definition of $u$ and $C$, and the second equality holds by~\eqref{eqC31}.

We now have that
\[
\begin{split}
\ff_{\loss}((C,u),x)&=\sum_{p \in C}u(p)(p-x)= \sum_{p \in C}u(p)p-x\sum_{p \in C}u(p)\\&= \sum_{p\in P}w(p)p-x\sum_{p\in P}w(p)=\ff_{\loss}((P,w),x),
\end{split}
\]
where the third equality is by~\eqref{eqC32}.

Hence, we obtain that $C' = (C,u)$ is an accurate coreset for $P' = (P,w)$, where $C$ is of size $|C| = d+2$, and its weight function $u$ is non-negative and sums to one over the points of $C$.

\paragraph{Generalized case of non-negative weights.}
We first remind the reader that $\displaystyle{\hat{p} = (p\mid 1)}$ for every $p\in P$  and $\hat{P} = \br{\hat{p}\mid p \in P}$. Since the Caratheodory Theorem cannot be applied to the weighted set $(P,w)$ whose weights do not necessarily sum to one, we apply the following steps: (i) define a new weighted set $(\hat{P},\hat{w})$ where  $\hat{w}:\hat{P} \to [0,1]$ such that $\displaystyle{\hat{w}(\hat{p}) = \frac{w(p) }{ \sum_{q\in P}w(q)}}$ for every $p\in P$, (ii) apply the Caratheodory Theorem to $(\hat{P},\hat{w})$ to compute a weighted set $(\hat{C},\hat{u})$ with the same weighted sum as $(\hat{P},\hat{w})$ as explained in Section~\ref{sec:cara}, and (iii) return the weighted set $(C,u)$ such that $C = \br{p \in P \mid \hat{p} \in \hat{C}}$ and $u:C\to [0,\sum_{p\in P}w(p)]$ where $u(p) = \hat{u}(\hat{p}) \cdot \sum_{p\in P}w(p) $.

Similarly to the proof in the previous case,  it is easy to verify that $u(p)\in [0,\sum_{p\in P} w(p)]$ for every $p\in C$, $\sum_{p\in P} w(p) = \sum_{p\in C} u(p)$, and for every $x\in \REAL^d$
\[
\ff_{\loss}((C,u),x)=\ff_{\loss}((P,w),x).
\]

See implementation of function \texttt{vectors\_sum\_3} in~\cite{opencode}.

\begin{figure}[h]
\begin{subfigure}[t]{0.46\textwidth}
    \includegraphics[width=\textwidth]{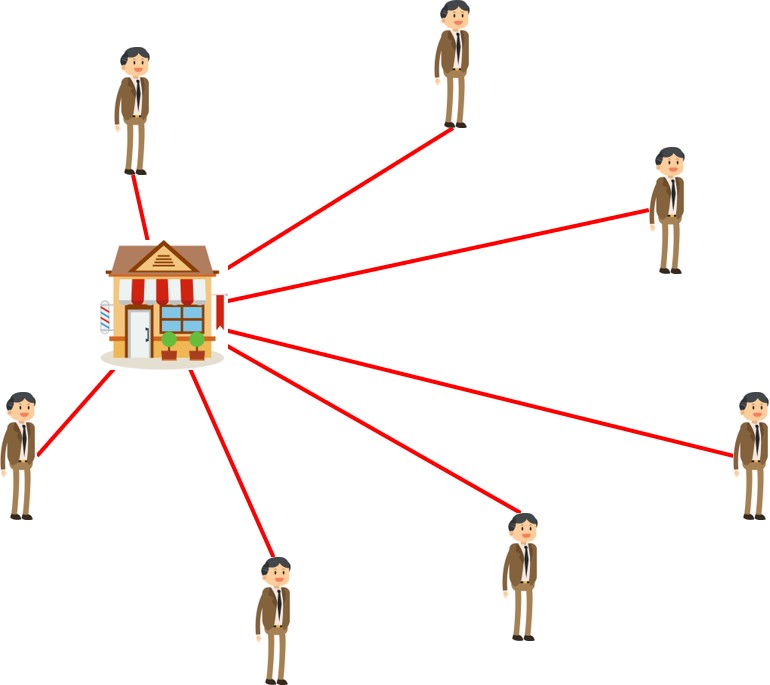}
    \caption{}
    \end{subfigure}
    ~~~~
    \begin{subfigure}[t]{0.46\textwidth}
    \includegraphics[width=\textwidth]{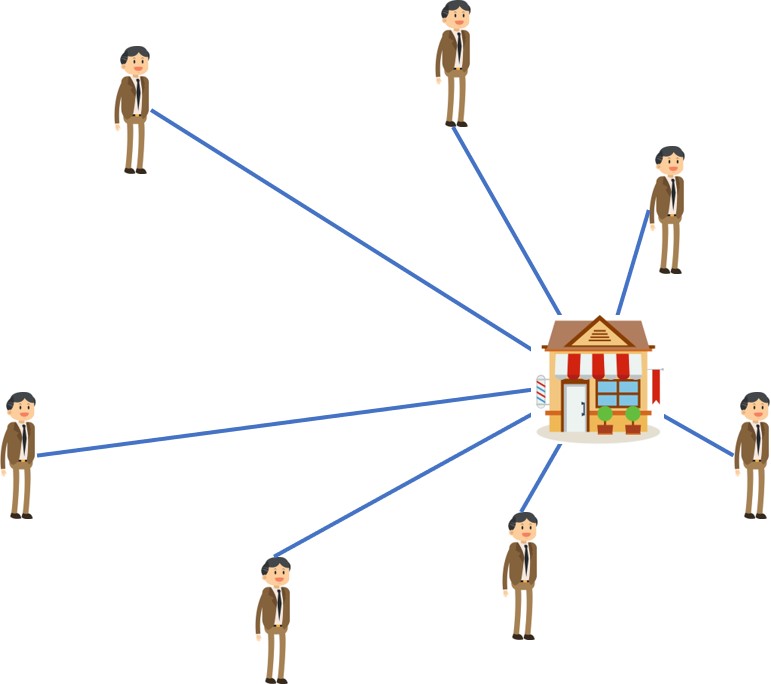}
    \caption{}
    \end{subfigure}
     \caption{\textbf{$1$-mean queries.} A set of resident locations, marked by the same humans in each of the two images. We are given $O(n)$ time to pre-process the resident locations. Then, given two potential locations for a shop as in the left or right images, we need to select the location that minimizes the average sum of squares distances (blue lines / red lines) in $O(1)$ time.}
    \label{fig:shopRMS}
\end{figure}

\subsection{$1$-Mean queries \medium} \label{1mean1}
Trying to minimize the distance from our shop to the farthest client is very sensitive to what is called outliers in the sense that a location of a single client $p$ (e.g. $p$ approaching infinity) may significantly change our cost function $\ff_{\loss}((P,w),x)$. A less sensitive cost function may wish to select the location that minimizes the average sum of squared (Least Mean Squared, LMS) distances from the clients to our shop. To this end, let $P\subseteq \REAL$ be a set of $|P|=n$ numbers, $\Y = \REAL$, $\ff(p,x) = (p-x)^2$, and $\loss(v) = \frac{1}{n}\norm{v}_1= \frac{1}{n} \sum_{i=1}^n v_i$ for every $v=(v_1,\cdots,v_n) \in \REAL^n$. The name $1$-mean queries or coreset was given since the mean of $P$ is the query $x\in\Y$ that minimizes this cost.
We wish to be able to compute the average sum of squared distances $\ff_{\loss}(P,x)=\frac{1}{n}\sum_{p\in P}(p-x)^2$ in $O(1)$ time to a (currently unknown) location $x\in\Y$ which will be given tomorrow, after pre-processing time of $O(n)$ today; See Fig.~\ref{fig:shopRMS}.

This can be done by observing that
\[
\ff_{\loss}(P,x)=\frac{1}{n}\sum_{p\in P}(p-x)^2=\frac{1}{n}\sum_{p\in P}\left(p^2-2xp+x^2\right)
=\frac{1}{n}\sum_{p\in P}p^2+x^2-2x\cdot \frac{1}{n}\sum_{p\in P}p.
\]
Hence, to evaluate the sum of squared distances from $P$ to $x$, all we need is to store the (coreset) $C=\br{\frac{1}{n}\sum_{p\in P}p^2, \frac{1}{n}\sum_{p\in P}p}$ which consists of two numbers. Clearly $C$ can be computed in $O(n)$ time and using $C$ we can compute $\ff_{\loss}(P,x)$ exactly for every number $x\in \Y = \REAL$ by defining a new function $\hh:\pow(\REAL)\times \REAL \to \REAL$ where $\hh(\br{a,b},x)=a+x^2-2xb$. Hence, $\hh(C,x)=\ff_{\loss}(P,x)$ for every $x\in\REAL$.

A similar solution holds for a set $P$ in $\REAL^d$ and $\Y = \REAL^d$ for any $d\geq1$ Euclidean space, and for any $w:P\to\REAL$ as
\[
\ff_{\loss}((P,w),x)=\sum_{p\in P}w(p)\norm{ p-x}^2=\sum_{p\in P}w(p)\norm{p}^2+ \norm{x}^2\sum_{p\in P}w(p)-2x^T\sum_{p\in P}w(p)p,
\]
where we assume that every $p\in P$ and $x\in\REAL^d$ is a column vector. Here, by letting
$C=\br{\sum_{p\in P}w(p)\norm{p}^2, \sum_{p\in P}w(p), \sum_{p\in P}w(p)p}$ contain two numbers and the weighted mean vector of $P$ and modifying $\hh$ as
\[
\hh(\br{a,b,c},x)=a + \norm{x}^2\cdot b - 2x^Tc
\]
for every $a,b\geq 0$ and $c\in\REAL^d$, we obtain $\ff_{\loss}((P,w),x)=\hh(C,x)$ for every $x\in \REAL^d$.
This set $C$ contains the second (variance), first (center of mass or mean), and zero moments of $P$.

Unlike in previous sections, in this example, this coreset $C$ is not a subset of the input data and we also use a new cost function.

\subsubsection{Subset coreset} \label{1mean2}
In this section we wish to construct a coreset for $1$-mean queries which uses the same cost function and is also a subset of the input.
For every $p\in P$, let $\hat{p}=(p^T \mid \norm{p}^2 \mid 1)^T$ be a corresponding vector in $\REAL^{d+2}$ and $\hat{P}=\br{\hat{p}\mid p\in P}$ be the union of these vectors.
Since the weighted mean $\sum_{\hat{p}\in \hat{P}}w(p)\hat{p}$ is spanned by $\hat{P}$ (i.e., linear combination of its subset), there is a subset $\hat{C} \subseteq \hat{P}$ of at most $|\hat{C}|=d+2$ points with a corresponding weight function $\hat{u}:\hat{C}\to\REAL$ such that
\begin{equation} \label{eqCmean}
\sum_{\hat{p}\in \hat{C}}\hat{u}(\hat{p})\hat{p} = \sum_{\hat{p}\in \hat{P}}w(p)\hat{p}.
\end{equation}
The set $\hat{C}$ can be computed as explained in Section~\ref{sec:LA}.

Let $C = \br{p\in P \mid \hat{p} \in \hat{C}}$, and let $u:C \to \REAL$ such that $u(p) = \hat{u}(\hat{p})$ for every $p\in C$.
We now have that
\begin{equation} \label{eqCms}
\begin{pmatrix}
\sum_{p\in C}u(p)p \\
\sum_{p\in C}u(p)\norm{p}^2\\
\sum_{p\in C}u(p)
\end{pmatrix} =
\sum_{\hat{p}\in \hat{C}}\hat{u}(\hat{p})\hat{p} = \sum_{\hat{p}\in \hat{P}}w(p)\hat{p} =
\begin{pmatrix}
\sum_{p\in P}w(p)p \\
\sum_{p\in P}w(p)\norm{p}^2\\
\sum_{p\in P}w(p)
\end{pmatrix},
\end{equation}
where the first equality is by the definitions of $C$ and $u$, the second equality is by~\eqref{eqCmean}, and the last equality is by the definition of $\hat{P}$.
Therefore, for every $x\in \REAL^d$, we have that
\[
\begin{split}
\ff_{\loss}((C,u),x) = \sum_{p\in C}u(p)\norm{ p-x}^2& =\sum_{p\in C}u(p)\norm{p}^2+ \norm{x}^2\cdot \sum_{p\in C}u(p)-2x^T\sum_{p\in C}u(p)p\\
& = \sum_{p\in P}w(p)\norm{p}^2+ \norm{x}^2\cdot \sum_{p\in P}w(p)-2x^T\sum_{p\in P}w(p)p\\
& = \ff_{\loss}((P,w),x),
\end{split}
\]
where the third derivation holds by~\eqref{eqCms}.


Unlike in the previous case, here the coreset is simply a scaled (weighted) subset of $P$ and the cost function $\ff_\loss$ is the same as for the input.

A natural question that comes to mind at this point is ``can we have a subset which is not weighted, i.e., $w(p)=1$ for every $p\in P$?''
Probably not, since this would imply that the mean of $P$ is the mean of a (non-weighted) subset of $P$ which cannot hold in general, even for a set $P$ of $3$ points on a line. Nevertheless, we can construct such a coreset that yields approximated answers to $1$-mean queries~\cite{inaba1994applications}. For the case of exact solution, we can still bound the weights as follows.

\subsubsection{Subset coreset of bounded weights} \label{1mean3}
In Section~\ref{1mean2} we used a linear combination of $d+2$ points from $\hat{P}$ to represent the weighted sum $\sum_{\hat{p}\in \hat{P}}w(p)\hat{p}$. Instead, if the input weights are non-negative, i.e., $w:P \to [0,\infty)$, we can apply Caratheodory's theorem, similarly to Section~\ref{shiftedMean3}, to compute a subset $\hat{C} \subseteq \hat{P}$ of size $|\hat{C}| = d+3$ along with a weights function $\hat{u}:\hat{C} \to [0,\sum_{p\in P}w(p)]$ that satisfies $\sum_{p\in\hat{C}}\hat{u}(p)=\sum_{p\in P}w(p)$ and
\[
\sum_{\hat{p}\in \hat{C}}\hat{u}(\hat{p})\hat{p} = \sum_{\hat{p}\in \hat{P}}w(p)\hat{p}.
\]

Now, by defining $C = \br{p\in P \mid \hat{p} \in \hat{C}}$ and $u(p) = \hat{u}(\hat{p})$ for every $p\in C$, we obtain for every $x\in\REAL^d$ that
\[
\ff_{\loss}((C,u),x) = \ff_{\loss}((P,w),x).
\]

See implementation of function \texttt{one\_mean\_3} in~\cite{opencode}.

\subsection{Coreset for $1$-segment queries \hard} \label{1segment}

\begin{figure}
  \centering
  \includegraphics[width=0.4\textwidth]{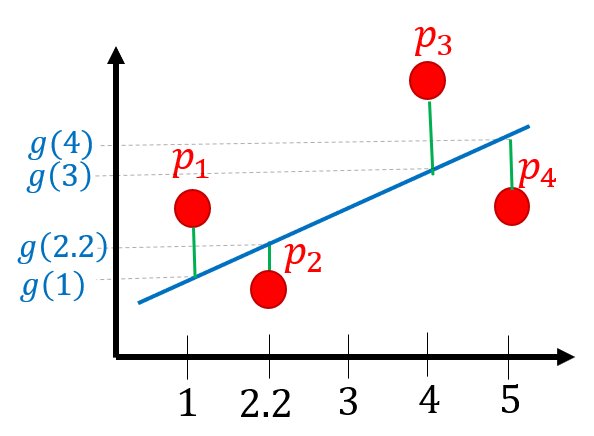}
  \caption{A signal $P = (t_1\mid p_1^T),\cdots,(t_4\mid p_4^T)$ (red dots), where $t_1 = 1, t_2 = 2.2, t_3 = 4,$ and $t_4=5$, and a $1$-segment $g:\REAL\to\REAL^2$ (blue line). The cost $\ff_{\loss}((P,\mathbf{1}),g)$ is the sum over the squared vertical
  distances (segments in greens) $\norm{p_i-g(t_i)}^2$ for every $i\in \br{1,\cdots,4}$.} \label{fig:1segmentIllustration}
\end{figure}

As stated in~\cite{rosman2014coresets}, there is an increasing demand for systems that learn long-term, high-dimensional data streams. Examples include video streams from wearable cameras, mobile sensors, GPS data, financial data, audio signals, and many more. In such data, a time instance is usually represented as a high-dimensional signal, for example location vectors, stock prices, or image content feature histograms. In other words, such data is usually represented as a set of linear segments.
Fast and real-time algorithms for summarization and segmentation of such large streams are of great importance, and can be made possible by compressing the input signals into a compact meaningful representation, which we call coreset for $1$-segment.

Let $(P,w)$ be a weighted set where $P = \br{(t_i \mid p_i^T)}_{i=1}^n \subseteq \REAL^{d+1}$ represents a (discrete) signal, for every $i\in [n]$, $t_i \in \REAL$ represents the time stamp and $p_i \in \REAL^d$ is a point, and $w:P\to [0,\infty)$. For simplicity we abuse notation and use $w(p)$ to denote $w((t\mid p^T))$ for every $(t \mid p^T) \in P$.  Let $\Y = \br{g \mid g:\REAL \to \REAL^d}$ be the set of all $1$-segments, $\ff\in \Y$ be a $1$-segment such that $\ff((t\mid p^T),g) = \norm{p-g(t)}^2$ for every $(t\mid p^T) \in P$ and $g \in \Y$, and $\loss(\cdot) = \norm{\cdot}_1$. Therefore, as shown in Fig.~\ref{fig:1segmentIllustration},
\[
\ff_{\loss}((P,w)g) = \sum_{(t\mid p^T)\in P}w(p) \norm{p-g(t)}^2.
\]

The goal is to compute a weighted set $(C,u)$ that represents a weighted (discrete) signal of size $|C| = d+2$, and $u:C \to [0,\infty)$ such that for every
$1$-segment $g\in \Y$, it holds that
\[
\ff_{\loss}((P,w),g) = c\cdot\ff_{\loss}((C,u),g).
\]

Put $g\in \Y$. Since $g$ is a linear segment, there exists $a,b\in \REAL^d$ such that $g(t) = a+b\cdot t$ for every $t\in \REAL$.

Let $X\in \REAL^{n\times (d+2)}$ be a matrix whose $i$th row is $\sqrt{w(p_i)} \cdot (1\mid t_i \mid p_i^T)$ for every $i\in [n]$. Let $U\Sigma V^T$ be the thin SVD of $X$; see Section~\ref{sec:LA}, and $u\in \REAL^{d+2}$ be the leftmost column of $\Sigma V^T$.

Let $c=\frac{\norm{u}^2}{d+2}$ and $Z \in \REAL^{(d+2)\times(d+2)}$ be an orthogonal matrix such that
\begin{equation} \label{eqYdef}
Zu = (\sqrt{c},\cdots,\sqrt{c})^T \in \REAL^{d+2},
\end{equation}
i.e., $Z$ can be regarded as a rotation matrix that rotates $u$ to the vector $(\sqrt{c},\cdots,\sqrt{c})^T$. Such a matrix $Z$ exists since $\norm{(\sqrt{c},\cdots,\sqrt{c})} = \norm{u}$.

Let $B\in \REAL^{(d+2)\times(d+1)}$ be the $(d+1)$ rightmost columns of $\frac{Z\Sigma V^T}{\sqrt{c}}$. Combining the definitions of $u$, $B$, and the fact that $Zu = (\sqrt{c},\cdots,\sqrt{c})^T$ yields that
\begin{equation} \label{eqYSV}
Z\Sigma V^T = \begin{bmatrix}
          \sqrt{c} &  \\
          \vdots & \sqrt{c} B \\
          \sqrt{c} &
        \end{bmatrix}.
\end{equation}

Let $C \subseteq \REAL^{d+2}$ be the union of rows of $B$ and $u:C\to [0,\infty)$ such that $u(p) = c$ for every $p\in C$. Then
\begin{align}
\ff_{\loss}((P,w),g) & = \sum_{(t\mid p^T)\in P} {w(p)}  \norm{p-g(t)}^2 = \sum_{(t\mid p^T)\in P} {w(p)}  \norm{ a+b\cdot t-p}^2\\
&= \sum_{(t\mid p^T)\in P}  \norm{ \sqrt {w(p)}(a+b\cdot t-p)}^2\nonumber
= \norm{\begin{bmatrix}
           \sqrt {w(p_1)}(1 \mid t_1 \mid p_1^T) \\
           \vdots \\
          \sqrt {w(p_n)}(1 \mid t_n \mid p_n^T)
        \end{bmatrix}
        \begin{bmatrix}
         a^T \\
         b^T \\
         -I
       \end{bmatrix}}^2 \nonumber\\
&=\label{eqSVD} \norm{U\Sigma V^T
        \begin{bmatrix}
         a^T \\
         b^T \\
         -I
       \end{bmatrix}}^2
= \norm{Z\Sigma V^T
        \begin{bmatrix}
         a^T \\
         b^T \\
         -I
       \end{bmatrix}}^2
= \norm{\begin{bmatrix}
          \sqrt{c} &  \\
          \vdots & \sqrt{c} B \\
          \sqrt{c} &
        \end{bmatrix}
        \begin{bmatrix}
         a^T \\
         b^T \\
         -I
       \end{bmatrix}}^2 \\
& = c\norm{\begin{bmatrix}
          1 &  \\
          \vdots & B \\
          1 &
        \end{bmatrix}
        \begin{bmatrix}
         a^T \\
         b^T \\
         -I
       \end{bmatrix}}^2
= \sum_{(t\mid p^T)\in B}c\norm{a+b\cdot t-p}^2 = \ff_{\loss}((C,u),g), \nonumber
\end{align}
where $(t\mid p^T)\in B$ denotes a row $(t \mid p^T)$ of $B$ which is the concatenation of a scalar $t\in \REAL$ and $p^T\in \REAL^d$, the first derivation in~\eqref{eqSVD} is by the definition of $U\Sigma V^T = X$, the second derivation in~\eqref{eqSVD} holds since $U$ and $Z$ are orthogonal matrices and the last derivation in~\eqref{eqSVD} is by~\eqref{eqYSV}. Therefore, the coreset $(C,u)$ is an accurate coreset for the given query space.

See implementation of function \texttt{one\_segment} in~\cite{opencode}.

\subsection{Coreset for Matrix $2$-norm \medium} \label{matrix2norm1}
A common approach to reduce the dimension of a high-dimensional data set $P$ in $\REAL^d$ is to project the vectors of $P$ (database records) onto some low $k$-dimensional affine subspace ($k$ is usually much smaller than $d$). For example, a subspace that minimizes the sum of squared distances ($\ell_2$ norm) to these input vectors, maybe under some constraints. Example algorithms include the Principle Component Analysis (PCA), Low-rank approximation ($k$-rank SVD) and Latent Dirichlet Analysis (LDA), and non-negative matrix factorization (NNMF).
Learning algorithms such as $k$-means clustering can then be applied on the low-dimensional data to obtain faster approximations with provable guarantees~\cite{feldman2013turning}. Dimensionality reduction is also used to avoid overfitting. Small number of features usually implies faster running/classification times and simpler models. Furthermore, smaller dimension means faster training of algorithms, less storage, less redundant features, and many more advantages.
However, the dimensionality reduction algorithms might be both time and space consuming. Therefore, to boost the running time of such algorithms, we can use accurate coresets as follows

Let $(P,w)$ be a weighted set where $P = \br{p_1,\cdots,p_n}$ is a set of $n$ points in $\REAL^d$ and $w:P\to [0,\infty)$ is a non-negative weights function, let $\Y = \REAL^d$, $x\in\Y$, $\ff(p,x)=(p^Tx)^2$ where $p^Tx=\langle p,x \rangle$ is the inner product of $p$ and $x$, for every $p\in \REAL^d$ and $x\in\Y$, and $\loss(\cdot) = \norm{\cdot}_1$. If we define $A$ as an $n\times d$ matrix whose $i$th row is $\sqrt{w(p_i)}p_i$, then
\[
\ff_{\loss}((P,w),x)=\norm{\left(w(p_1)(p_1^Tx)^2,\cdots,w(p_n)(p_n^Tx)^2\right)}_1=\sum_{p\in P}w(p)(p^Tx)^2=\norm{Ax}^2.
\]

\begin{figure}[h]
	\begin{subfigure}[t]{0.5\textwidth}
    \includegraphics[width=\textwidth]{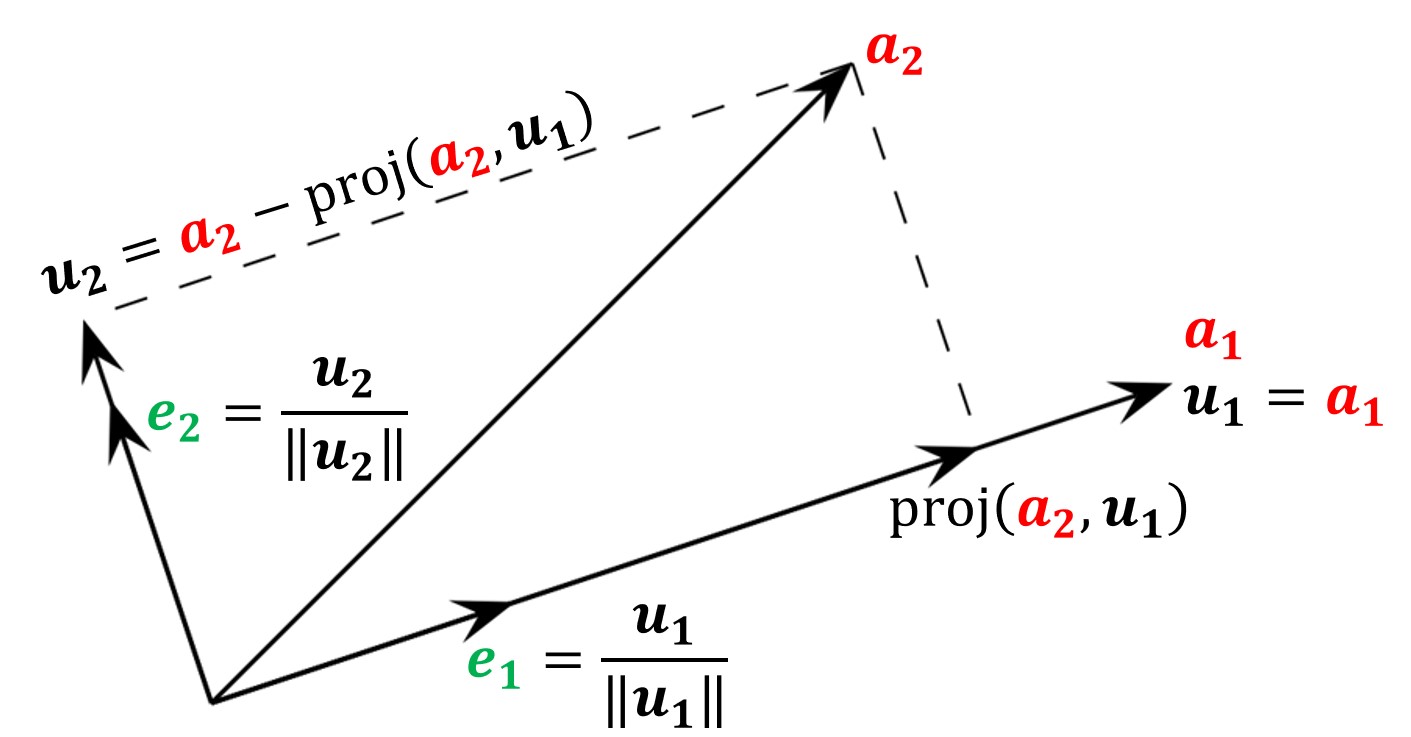}
    \end{subfigure}
    ~
    \begin{subfigure}[t]{0.5\textwidth}
    \includegraphics[width=\textwidth]{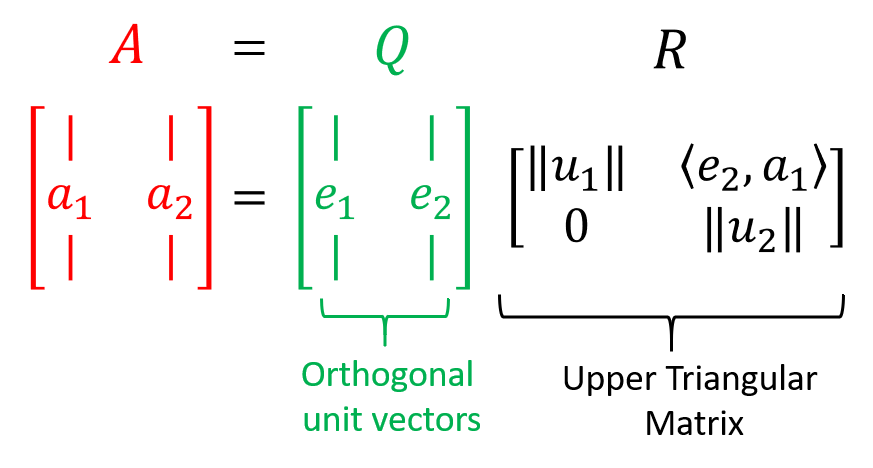}
    \end{subfigure}
        \caption{\textbf{$QR$-Decomposition.} \textbf{Left:} Given a set of linearly independant vectors $\br{a_1,a_2}\subseteq \REAL^2$. The goal is generate an orthogonal set $\br{e_1,e_2} \subseteq \REAL^2$ that spans the same subspace as the given set. This process is also called the Gram-Schmidt algorithm in Linear Algebra. \textbf{Right:} The result is a QR-decomposition of the matrix $A$ whose columns are ${a_1,a_2}$ into an orthogonal matrix $Q$ (i.e., $Q^TQ=I_2$) and an upper triangular matrix $R$.}
    \label{fig:QRDecomp}
\end{figure}

\begin{figure}[h]
	\begin{subfigure}[t]{0.5\textwidth}
\centering
    \includegraphics[width=0.6\textwidth]{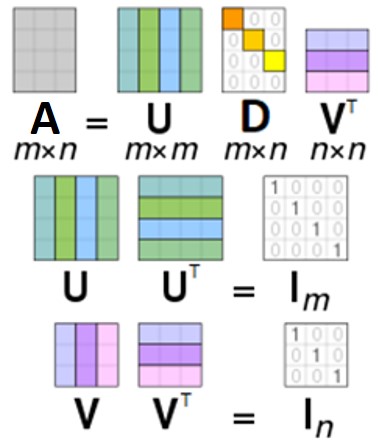}
    \end{subfigure}
    ~
    \begin{subfigure}[t]{0.5\textwidth}
    \centering
    \includegraphics[width=0.7\textwidth]{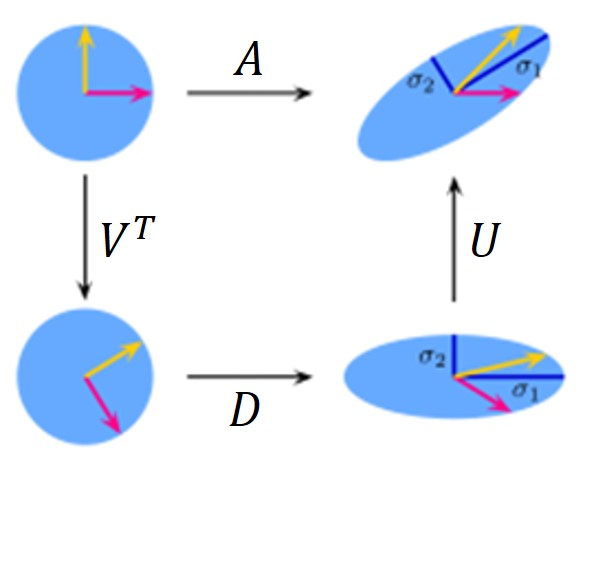}
    \end{subfigure}
        \caption{\textbf{Singular Value Decomposition.} \textbf{Left:} A Singular Value Decomposition $UDV^T$ of a matrix $A \in \REAL^{m\times n}$, where $U \in \REAL^{m\times m}$ and $V\in \REAL^{n\times n}$ are orthogonal matrices and $D\in \REAL^{m\times n}$ is a diagonal matrix that contains the singular values of $A$.
        \textbf{Right:} Visualization of the SVD of a matrix $A \in \REAL^{2\times 2}$. The matrix $M$ distorts the unit disc to an ellipse. The SVD is a decomposition of $A$ into three simple transformations: an initial rotation (possibly with reflection) $V^T$, a scaling $D$ along the coordinate axes, and a final rotation (possibly with reflection) $U$. The lengths $\sigma_1$ and $\sigma_2$ of the semi-axes of the ellipse are the singular values of $A$, namely $d_1$ and $d_2$. Illustrations taken from~\protect\cite{wikiSVD}.}
    \label{fig:SVD}
\end{figure}

\begin{figure}
  \centering
  \includegraphics[width=0.5\textwidth]{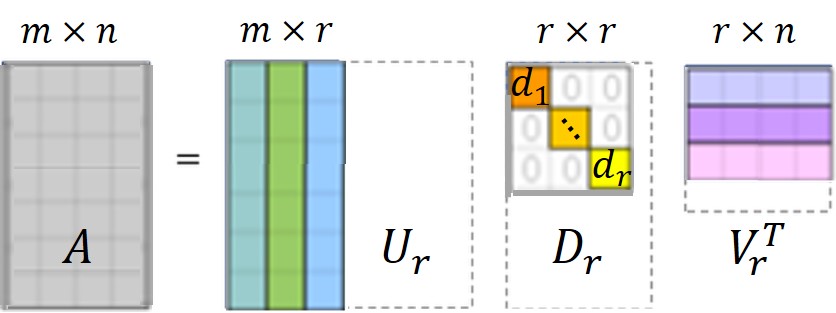}
  \caption{\textbf{Thin Singular Value Decomposition. }A Thin Singular Value Decomposition (thin SVD) $U_r D_r V_r^T$ of a matrix $A \in \REAL^{m\times n}$ of rank $r \leq n$, where $U_r \in \REAL^{m\times r}$, $V_r \in \REAL^{n\times r}$, $U_r^TU_r = I_r$, $V_r^TV_r = I_r$, and $D \in \REAL^{r\times r}$ is a diagonal matrix containing the singular values $d_1 \geq \cdots \geq d_r > 0$.}
  \label{fig:thinSVD}
\end{figure}

We aim to compute a matrix $\mathcal{C}\in\REAL^{d\times d}$ such that $\norm{Ax}^2=\norm{\mathcal{C}x}^2$.
This can be done by letting $Q\in\REAL^{n\times d}$ be a matrix of orthogonal columns ($Q^TQ=I$) that spans the $d$ columns of $A$, e.g. via Grahm-Shmidt (also known as the $A=QR$ as shown in Fig.~\ref{fig:QRDecomp}), or the Thin Singular Value Decomposition ($A=U_rD_rV_r^T$ as shown in Fig.~\ref{fig:SVD}--\ref{fig:thinSVD}). By letting $\mathcal{C}=R$ be a $d\times d$ matrix whose columns correspond to the columns of $A$ under the base $Q$, we obtain $A=QR = Q\mathcal{C}$. Since $Q$ has orthogonal columns, we have $\norm{Qx}=\norm{x}$ for every $x\in\REAL^d$. Therefore, by defining $C \subseteq \REAL^d$ to contain the rows of $\mathcal{C}$ we obtain that
\begin{equation}\label{CC}
\ff_{\loss}((P,w),x)=\norm{Ax}^2=\norm{Q\mathcal{C}x}^2=\norm{\mathcal{C}x}^2=\ff_{\loss}((C,\mathbf{1}),x).
\end{equation}
Note that, without loss of generality, we can assume that $\norm{x}=1$, i.e., that $\Y$ is a set of only unit vectors. This is because for every vector $y\in\REAL^d$ there is $c\geq 0$ and a unit vector $x=y/\norm{y}$ such that
\[
\ff_{\loss}((P,w),y)=\norm{Ay}^2=\norm{Acx}^2=c^2\norm{Ax}^2=c^2\norm{\mathcal{C}x}^2=\norm{\mathcal{C}cx}^2=\norm{\mathcal{C}y}^2=\ff_{\loss}((C,\mathbf{1}),y).
\]

Geometrically, if $x$ is a unit vector, $\ff(p,x)=(p^Tx)^2$ is the squared distance between a point $p$ in $\REAL^d$ and a hyperplane that intersects the origin and is orthogonal to $x$. More generally, the coreset $(C,\mathbf{1})$ from~\eqref{CC} can be used to compute the weighted sum of squared distances $\dist^2((P,w),S)$ over the points of $P$ to a $j$-dimensional subspace of $\REAL^d$ that is spanned by the orthonormal columns of a matrix $S\in\REAL^{d\times j}$, i.e., $S^TS=I$. Let $S^\bot\in\REAL^{d\times (d-j)}$ denote the matrix whose columns span the subspace that is orthogonal to $S$, i.e., $[S \mid S^\bot]^T[S \mid S^\bot] = I$. Observe that by the definition of $S^\bot$ we have $\dist^2((P,w),S)=\norm{A{S^\bot}}_F^2$.
Then, by the Pythagorean Theorem
\begin{equation} \label{genToJSub}
\dist^2((P,w),S)
=\norm{A{S^\bot}}_F^2
=\sum_{i=1}^{d-j} \norm{A{S^\bot}_{*i}}^2
=\sum_{i=1}^{d-j} \norm{\mathcal{C}{S^\bot}_{*i}}^2
=\dist^2((C,\mathbf{1}),S),
\end{equation}
where ${S^\bot}_{*i}$ denotes the $i$th column of ${S^\bot}$, and where the third equality is by the coreset property in~\eqref{CC}.


\subsubsection{Subset coreset of bounded weights} \label{matrix2norm2}
Recall that $A\in \REAL^{n\times d}$ is a matrix whose rows are the weighted points $\br{\sqrt{w(p_i)}p_i}_{i=1}^n$. The set $C \subseteq \REAL^{d}$ from the previous example contains $d$ points in $\REAL^d$ but are not a subset of the input set $P$. This has few disadvantages that were discussed in Section~\ref{sec:intro}.
We now show how to obtain such a weighted set $(C,u)$ where $C$ is a subset of $P$, where every point $p$ in $C$ is also assigned to a non-negative weight $u(p)\in[0,\infty)$. Observe that for every $x\in \REAL^d$,
\[
\ff_{\loss}((P,w),x)= \norm{Ax}^2={x}^TA^TAx={x}^T\left(\sum_{p\in P}w(p)pp^T\right){x}.
\]

For every $p \in P$, the $d\times d$ matrix $pp^T$ corresponds to a vector $\hat{p}\in \REAL^{d^2}$ by concatenating its rows. Hence, $\sum_{p\in P}w(p)\tovec(pp^T)=\sum_{p\in P}w(p)\hat{p}$, where $\tovec(M) \in \REAL^{d^2}$ is a row stacking of the matrix $M\in\REAL^{d\times d}$. Let $\hat{P} = \br{\hat{p}\mid p\in P}$ and let $\hat{w}(\hat{p}) = \frac{w(p)}{\sum_{q\in P}w(q)}$ for every $\hat{p}\in \hat{P}$.
Since $\hat{P} \subseteq \REAL^{d^2}$ and the weighted sum of $\left(\hat{P},\hat{w}\right)$ lies inside the convex hull of $\hat{P}$, by applying Caratheodory's Theorem on the weighted set $\left(\hat{P},\hat{w}\right)$, we obtain that there is a subset $\hat{C}\subseteq \hat{P}$ of size $|\hat{C}|=d^2+1$ and a weights function $\hat{u}:\hat{C}\to [0,1]$ such that
\begin{equation} \label{eq:mulSumw}
\frac{1}{\sum_{p\in P}w(p)}\sum_{\hat{p}\in \hat{P}}w(p)\hat{p}= \sum_{\hat{p}\in \hat{P}}\hat{w}(\hat{p})\hat{p} = \sum_{\hat{p}\in \hat{C}}\hat{u}(\hat{p})\hat{p}.
\end{equation}
Multiplying~\eqref{eq:mulSumw} by $\sum_{p\in P}w(p)$ we obtain
\begin{equation} \label{caraCoresetd2}
\sum_{\hat{p}\in \hat{P}}w(p)\hat{p}=\sum_{p\in P}w(p) \cdot \sum_{\hat{p}\in \hat{C}}\hat{u}(\hat{p})\hat{p}.
\end{equation}

Let $C = \br{p \in P \mid \hat{p} \in \hat{C}}$ and let $u:C\to [0,\sum_{p\in P}w(p)]$ such that $u(p) = \displaystyle \hat{u}(\hat{p})\cdot \sum_{p\in P}w(p)$ for every $p\in C$.
Combining the definitions of $C$, $u$ and~\eqref{caraCoresetd2} yields
\[
\sum_{p\in P}w(p)pp^T= \sum_{p\in P}w(p) \cdot \sum_{p\in C}\hat{u}(\hat{p})pp^T = \sum_{p\in C}u(p)pp^T.
\]
Hence, by letting $Z = \left(\sqrt{u(p)}\cdot p\right)_{p\in C}^T \in \REAL^{(d^2+1)\times d}$ be a matrix whose rows $\sqrt{u(p)} \cdot p^T$ are weighted (scaled) points of $C$, we obtain that for every $x\in \REAL^d$
\begin{equation} \label{caraCoresetMatrices}
\begin{split}
\ff_{\loss}((P,w),x)
&= \norm{A{x}}^2
={x}^TA^TA{x}
={x}^T\left(\sum_{p\in P}w(p)pp^T\right){x}\\
& ={x}^T\left(\sum_{p\in C}u(p)pp^T\right){x}
= {x}^TZ^TZ{x} = \norm{Zx}^2
= \ff_{\loss}((C,u),x).
\end{split}
\end{equation}
Therefore, $(C,u)$ is an accurate coreset for the query space $(P,w,\REAL^d,\ff,\norm{\cdot}_1)$ where $\ff(p,x) = (p^Tx)^2$.

Similarly to~\eqref{genToJSub} in Section~\ref{matrix2norm1}, for every $j$-dimensional subspace of $\REAL^d$ that is spanned by the orthonormal columns of a matrix $S\in\REAL^{d\times j}$, and its orthogonal complement $S^\bot$, we have that
\[
\dist^2((P,w),S)
=\norm{A{S^\bot}}_F^2
=\sum_{i=1}^{d-j} \norm{A{S^\bot}_{*i}}^2
=\sum_{i=1}^{d-j} \norm{Z{S^\bot}_{*i}}^2
=\dist^2((C,u),S),
\]
where the third derivation holds by~\eqref{caraCoresetMatrices}.

See implementation of function \texttt{matrix\_norm2} in~\cite{opencode}.

\subsection{Least-Mean-Squares Solvers \hard} \label{sec:LMS}
Least-mean-squares solvers are very common optimization methods in machine learning and statistics. They are typically used for normalization, spectral clustering, feature selection, prediction, classification, and many more. In this section, we define and derive a coreset for such problems, based on the coreset presented in Section~\ref{matrix2norm2}.

The corresponding query space $(P,w,\REAL^d,\ff,\loss)$ for least mean squares problems is as follows.
Let $(P,w)$ be a weighted set where $P = \br{(a_1^T\mid b_1),\cdots,(a_n^T \mid b_n)} \subseteq \REAL^{d+1}$, $w:P\to [0,\infty)$, and for every $i\in [n]$, $a_i \in \REAL^d$ and $b_i \in \REAL$. Let $\Y = \REAL^d$,  and for every $(a^T\mid b)^T \in \REAL^{d+1}$ where $a \in \REAL^d$ and $b \in \REAL$ let
$\ff((a^T\mid b),x) = (a^Tx-b)^2$, and define $\loss = \norm{\cdot}_1$.
For simplicity let  $w_i = w((a_i^T\mid b_i))$ for every $i\in [n]$.  Therefore,
\begin{equation} \label{eq:LMSLoss}
\ff_{\loss}((P,w),x) = \sum_{i=1}^n  w_i(a_i^Tx-b_i)^2.
\end{equation}

To obtain an accurate coreset for the above query space $(P,w,\REAL^d,\ff,\loss)$, we shall define a new and slightly different query space, and use the accurate coreset from Section~\ref{matrix2norm2} as follows.

Let $f_2(p,x) = (p^Tx)^2$ for every $p,x\in \REAL^{d+1}$. Let $(C,u)$ be an accurate coreset of size $|C|=(d+1)^2 +1$ for the new query space $(P,w,\REAL^{d+1},f_2,\norm{.}_1)$ as explained in Section~\ref{matrix2norm2}, where $C = \br{(\hat{a}_1^T\mid \hat{b}_1),\cdots, (\hat{a}_{{|C|}}^T \mid \hat{b}_{|C|})} \subseteq P$. For simplicity define  $u_i = u((\hat{a}_i^T,\hat{b}_i))$ for every $i\in [|C|]$. Then for every $x'\in \REAL^{d+1}$,
\begin{align*}
\sum_{i=1}^n w_i((a_i^T\mid b_i)x')^2
=\sum_{i=1}^{|C|} u_i( (\hat{a}^T_i\mid \hat{b}_i)x')^2.
\end{align*}

Since the last equality holds for every $x'\in \REAL^{d+1}$, in particular, for every $x'=(x^T\mid -1)^T \in \REAL^{d+1}$ where $x\in \REAL^d$, we have that :
\begin{align}
\ff_{\loss}((P,w),x)
&=\sum_{i=1}^n w_i(a_i^Tx-b)^2
= \sum_{i=1}^n w_i((a_i^T\mid b_i)x')^2\\
&= \sum_{i=1}^{|C|} u_i((\hat{a}^T_i\mid \hat{b_i})x')^2
= \sum_{i=1}^{|C|} u_i(\hat{a}_i^Tx-\hat{b})^2
=\ff_{\loss}((C,u),x).
\end{align}
Hence, $(C,u)$ is an accurate coreset for the query space $(P,w,\REAL^d,\ff,\loss)$.

Commonly in the field of machine learning, least mean squares optimization problems are defined using matrix notations as follows. Let  $A = (\sqrt{w_1}a_1 \mid \cdots \mid \sqrt{w_n} a_n)^T \in \REAL^{n\times d}$ and $b = (\sqrt{w_1}b_1,\cdots,\sqrt{w_n}b_n)^T \in \REAL^n$.
Least-mean-squares solvers typically aim to minimize
\begin{equation}\label{eqAxb}
 g\left( \sum_{i=1}^n  w_i(a_i^Tx-b_i)^2\right)+h(x) =  g\left(\norm{Ax-b}_2^2\right)+h(x)
\end{equation}
over every $x\in \Y \subseteq \REAL^d$. Here, $h:\REAL^d \to [0,\infty)$ is called a \emph{regularization function} on the parameters of $x$, and it is independent of $(P,w)$, and $g:\REAL \to \REAL$ is a real function. See Table~\ref{table:examples} for example objective functions.

\begin{table}[h!]
\centering
\begin{adjustbox}{width=\textwidth}
\begin{tabular}{|c|c|c|c|}
\hline & \\[-10pt]
\bf Solver name  &  \bf Objective function & $g(x)$ & $h(x)$\\[2pt]
\hhline{|=|=|=|=|} &  \\[-14pt]
Linear regression~\cite{bjorck1967solving}  & $\displaystyle {\norm{Ax-b}_2^2}$ & $x$ & $0$ \\[15pt]
\hline & \\[-14pt]
Ridge regression~\cite{hoerl1970ridge}  & $\displaystyle{{\norm{Ax-b}_2^2}+\alpha\norm{x}_2^2}$ & $x$ & $\alpha \norm{x}_2^2$  \\[15pt]
\hline& \\[-14pt]
Lasso regression~\cite{tibshirani1996regression}  & $\displaystyle{\frac{1}{2n}{\norm{Ax-b}_2^2}+\alpha\norm{x}_1}$ & $\frac{x}{2n}$ & $\alpha \norm{x}_1$ \\[15pt]
\hline &\\[-14pt]
Elastic-Net regression~\cite{zou2005regularization}  &$\displaystyle\frac{1}{2n}{\norm{Ax-b}_2^2}+ \rho\alpha\norm{x}_2^2 +\frac{(1-\rho)}{2}\alpha\norm{x}_1$ & $\frac{x}{2n}$ & $\rho\alpha\norm{x}_2^2 +\frac{(1-\rho)}{2}\alpha\norm{x}_1$ \\[15pt]
\hline
\end{tabular}
\end{adjustbox}\vspace{0.2cm}\caption{Example solvers that aim to minimize objective functions in the form of~\eqref{eqAxb}. Each solver gets a matrix $A\in\REAL^{n\times d}$, a vector $b\in\REAL^n$ and aims to compute $x\in\REAL^d$ that minimizes the objective function. Additional given regularization parameters include $\alpha>0$ and $\rho\in [0,1]$.}\label{table:examples}

\end{table}

Hence, as we defined the matrix $A$ and the vector $b$ based on the weighed set $(P,w)$, we define the matrix $Z$  and the vector $v$ based on the coreset $(C,u)$ of the query space $(P,w,\REAL^d,\ff,\loss)$,
i.e., $Z \in \REAL^{ m\times d }$ such that the $i$th row of $Z$ is $\sqrt{u_i} \hat{a}_i^T$, and $v\in \REAL^{m } $ such that $v= (\sqrt{u_1}\hat{b}_1, \cdots, \sqrt{u_{|C|}}\hat{b}_{|C|} )$ and $m= (d+1)^2 +1$.

Now as desired, the family of functions from~\eqref{eqAxb} satisfy that,
\begin{align*}
g(\norm{Ax-b}_2^2)+h(x)
&= g\left( \sum_{i=1}^n  w_i(a_i^Tx-b_i)^2\right)+h(x) \\
&=g\left( \sum_{i=1}^m  w_i(\hat{a}_i^Tx-\hat{b}_i)^2\right)+h(x)
=g(\norm{Zx-v}_2^2)+h(x).
\end{align*}

See implementation of function \texttt{LMS\_solvers} in~\cite{opencode}.

%
%
%
%

\section{Caratheodory's Theorem} \label{sec:cara}
The Caratheodory Theorem~\cite{caratheodory1907variabilitatsbereich,cook1972caratheodory} is a fundamental result in computational geometry that states that if a point $x\in \REAL^d$ lies inside the convex hull of a set $P \subseteq \REAL^d$, of $|P|=n$ points i.e., $x$ is a convex combination of the points of $P$, then there is a subset of at most $d+1$ points from $P$ that contains $x$ in its convex hull, i.e., $x$ can be represented as a convex combination of at most $d+1$ points from $P$; see Theorem~\ref{CaraTheorem}.

\begin{figure}[hbtp]
\centering
\includegraphics[width=0.6\textwidth]{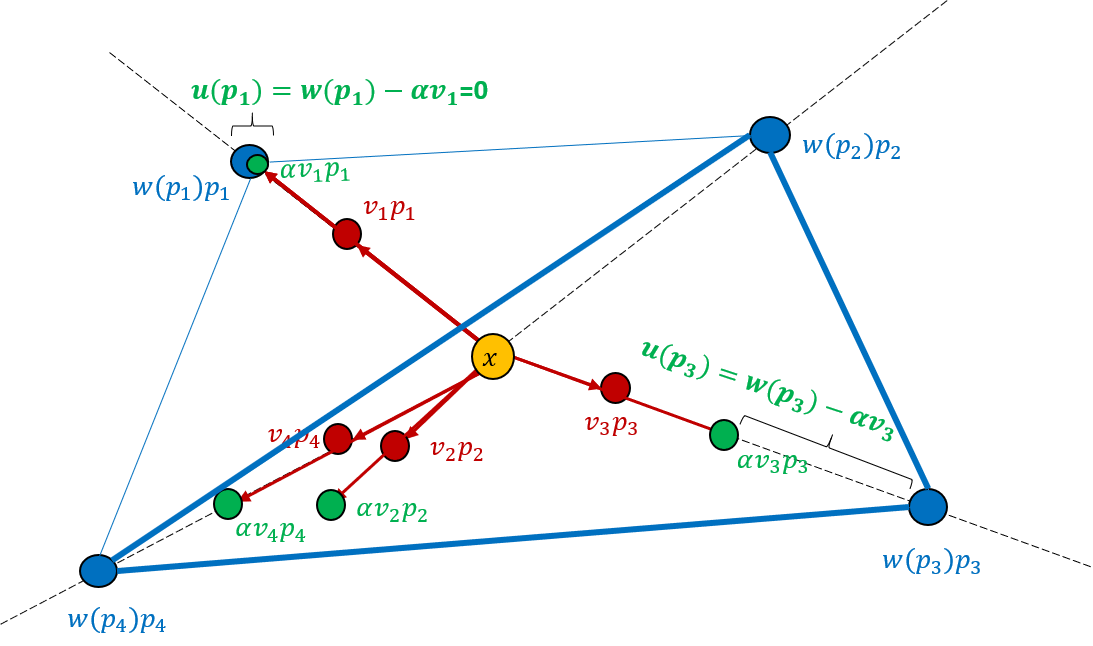}
\caption{\scriptsize A weighted set $(P,w)$ whose weighted sum is $\sum_{i=1}^4 w(p_i)p_i=0$ corresponds to four points (in blue) whose weighted sum is the point $x$ (in orange), which we assume is the origin $x=\vec{0}$. Algorithm~\ref{Alg:cara} first computes a weights vector $v=(v_1,\cdots,v_4)^T$ such that the weighted sum $\sum_{i=1}^4 v_ip_i$ (red points) is the origin, and sum of weights is $\sum_{i=1}^4 v_i=0$. The weights are scaled by $\alpha>0$ until $\alpha v_i=w(p_i)$ for some $i$ ($i=1$ in the figure).
Every point in the resulting set $\br{\alpha v_i p_i}_{i=1}^n$ in green is subtracted from its corresponding point in the input set $\br{w(p_i) p_i}_{i=1}^n$ to obtain the output set $\br{u(p_i)p_i}_{i=1}^n = \br{(w(p_i)-\alpha v_i)p_i}_{i=1}^n$, where $u(p_1)=0$ so $p_1$ can be removed. Algorithm~\ref{Alg:cara} then continues iteratively with the remaining points until $(P,u)$ has $|P|=d+1$ weighted points. The figure is taken from~\protect\cite{nasser2015coresets}.}
\label{fig:caraProof}
\end{figure}

The proof of Caratheodory's theorem is constructive, and the above subset of $d+1$ points can be computed in $O(n^2d^2)$ time; see Algorithm~\ref{Alg:cara} which implements this constructive proof. An intuition behind the proof of correctness is shown in Fig.~\ref{fig:caraProof}.
The algorithm takes as input a weighted set $(P,w)$ such that $w:P\to [0,1]$ and $\sum_{p\in P}w(p) = 1$, and computes in $O(n^2d^2)$ time a new weighted set $(S,u)$ such that $S \subseteq P$, $|S| \leq d+1$, $u:s\to [0,1]$, $\sum_{s\in S} u(s) = 1$ and $\sum_{s\in S} u(s)s = \sum_{p\in P}w(p) p$.

\begin{theorem} \label{CaraTheorem}
If a point $x\in \REAL^d$ is in the convex hull of a set $P \subseteq \REAL^d$, then $x$ is also in the convex hull of a set of at most $d+1$ points from $P$.
\end{theorem}

\setcounter{AlgoLine}{0}
\begin{algorithm}[th]
\caption{$\caras(P,w)$}\label{Alg:cara}
\SetKwInOut{Input}{Input}
\SetKwInOut{Output}{Output}
\Input{A weighted set $(P,w)$ of $n$ points in $\REAL^{d}$, where $P = \br{p_1,\cdots,p_n}$, $w:P\to [0,1]$ and $\sum_{p\in P} w(p) = 1$.}
\Output{A weighted set $(S,u)$ computed in $O(n^2d^2)$ time such that $S\subseteq P$, $|S| \leq d+1$, $u:S\to[0,1]$, $\sum_{s\in S} u(s) = 1$ and $\sum_{s\in S} u(s)s = \sum_{p\in P}w(p) p$.}

\If{$n\leq d+1$}
{\Return $(P,w)$}
\For {every $i\in\br{2,\cdots,n}$}
{
$a_i := p_i - p_1$ \tcp{$p_i$ is the $i$th point of $P$.}
}

$A := ( a_2 \mid \cdots \mid a_{n})$ \tcp{$A\in \REAL^{d\times (n-1)}$}

Compute $v=(v_2,\cdots,v_{n})^T\neq 0$ such that $Av=0$. \label{l11}\\

$\displaystyle v_1 := -\sum_{i=2}^{n}  v_i$\label{u1}\\

$\displaystyle \alpha := \min\br{\frac{w(p_i)}{v_i} \mid i\in \br{1,\cdots,n} \text{ and }  v_i> 0}$ \label{alp}\\

$u(p_i) := (w(p_i)-\alpha v_i)$ for every $i\in\br{1,\cdots,n}$ such that $u(p_i)>0$. \label{Sdef}\\

$S := \br{p_i\mid i\in\br{1,\cdots,n} \text{ and } u(p_i)>0}$\label{Sdeff}\\
\If {$|S|>d+1$}{
$(S,u):= \caras(S,u)$ \label{eight} \tcp{Recursive call that reduces $S$ by at least $1$.}
}
\Return $(S,u)$
\end{algorithm}

\textbf{The proof of correctness for Theorem~\ref{CaraTheorem}.}
The proof of correctness for Theorem~\ref{CaraTheorem} follows from the correctness of the procedure $\caras$ and vice versa; see Algorithm~\ref{Alg:cara}.
The procedure $\caras$ takes as input a weighted set $(P,w)$ whose points are denoted by $P=\br{p_1,\cdots,p_n}$ and where $\sum_{p\in P} w(p) = 1$. We assume $n > d+1$, otherwise $(S,u)=(P,w)$ is the desired output. Hence, the $n-1 > d$ points $p_2-p_1$, $p_3-p_1, \ldots, p_n-p_1$ must be linearly dependent. This implies that there are reals $v_2,\cdots,v_n$, which are not all zeros, such that
\begin{equation}\label{eq00}
\sum_{i=2}^n v_i (p_i-p_1)=0.
\end{equation}
These reals are computed in Line~\ref{l11} by solving a system of linear equations. This step dominates the running time of the algorithm and takes $O(nd^2)$ time using e.g. SVD, where the desired vector of coefficients $(v_2,\cdots,v_n)^T$ is simply the right singular vector that corresponds to the smallest singular value in the SVD of the matrix $M = (p_2-p_1, \cdots, p_n-p_1)^T \in \REAL^{(n-1)\times d}$.

The definition
\begin{equation}\label{uudef}
v_1=-\sum_{i=2}^n v_i
\end{equation}
in Line~\ref{u1}, guarantees that
\begin{equation} \label{negativev}
v_j < 0 \text{ for some }j\in [n],
\end{equation}
and that
\begin{align}
\sum_{i=1}^n v_i p_i&=v_1p_1+\sum_{i=2}^n v_i p_i \nonumber\\
& =\left(-\sum_{i=2}^n v_i\right) p_1+\sum_{i=2}^n v_i p_i \label{eqq2}\\
& =\sum_{i=2}^n v_i (p_i-p_1)=0, \label{eqq}
\end{align}
where~\eqref{eqq2} is by~\eqref{uudef} and the second equality in~\eqref{eqq} is by~\eqref{eq00}.
Hence, for every $\alpha\in\REAL$, the weighted sum of $P$ is
\begin{equation}\label{sum}
\sum_{i=1}^n w(p_i)p_i=\sum_{i=1}^n w(p_i)p_i-\alpha\sum_{i=1}^n v_i p_i=  \sum_{i=1}^n \left(w(p_i)-\alpha v_i\right) p_i,
\end{equation}
where the first equality holds since $\sum_{i=1}^n v_i p_i = 0$ by~\eqref{eqq}.
The definition of $\alpha$ in Line~\ref{alp} guarantees that $\alpha v_{i^*}=w(p_{i^*})$ for some $i^*\in[n]$, and that $w(p_i)-\alpha v_i\geq 0$ for every $i\in[n]$. Hence, the set $S$ that is defined in Line~\ref{Sdeff} contains at most $n-1$ points, its weighted sum is equal to the weighted sum of $(P,w)$, and its set of weights $\br{w(p_i)-\alpha v_i}$ is non-negative.

Notice that if $\alpha = 0$, we have that $u(p_k) = w(p_k) > 0$ for some $k\in [n]$. Otherwise, by~\eqref{negativev}, there is $j\in [n]$ such that $u(p_j) = w(p_j) -\alpha v_j > 0$. Hence, $|S| \neq \emptyset$.
The sum of the positive weights is thus equal to the sum of input weights,
\[
\sum_{p_i\in S}^n u(p_i)=\sum_{i=1}^n (w(p_i)-\alpha v_i)=\sum_{i=1}^nw(p_i)-\alpha\cdot \sum_{i=1}^n v_i=1,
\]
where the last equality holds by~\eqref{uudef} and since the sum of the input weights is $1$, i.e., $\sum_{i=1}^nw(p_i) = 1$. This and~\eqref{sum} proves the desired properties of $(S,u)$, which is of size $n-1$. In Line~\ref{eight} we repeat this process recursively until there are at most $d+1$ points left in $S$. For $O(n)$ iterations the overall time is thus $O(n^2d^2)$.

\subsection{Faster Construction}

\begin{figure}
  \centering
  \includegraphics[width=\textwidth]{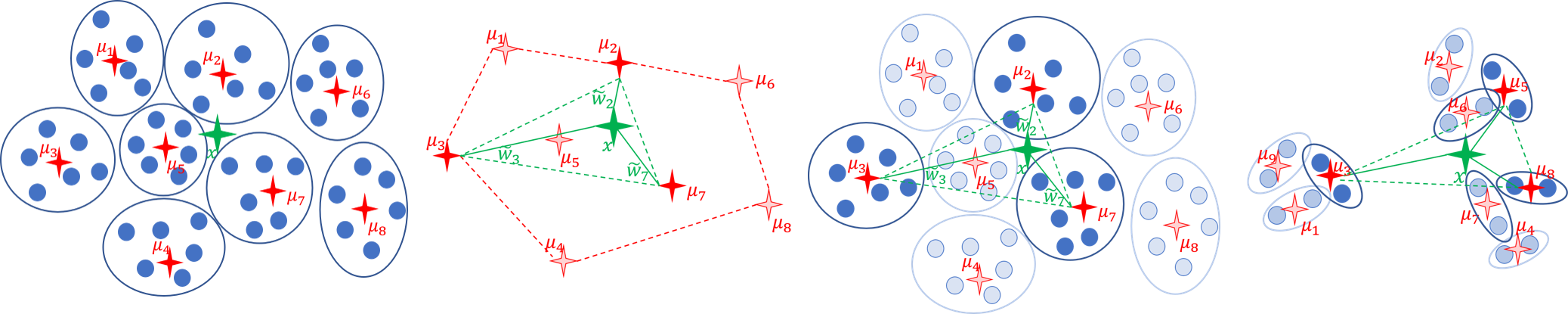}
  \caption{\scriptsize Overview of the improved Caratheodory algorithm in~\protect\cite{maalouf2019fast}. Images left to right: Steps (i) and (ii): A partition of the input weighted set of $n=48$ points (in blue) into $k=8$ equal clusters (in circles) whose corresponding means are $\mu_,\ldots,\mu_8$ (in red). The mean of $P$ (and these means) is $x$ (in green). Step (iii): Caratheodory (sub)set of $d+1=3$ points (bold red) with corresponding weights (in green) is computed only for these $k=8\ll n$ means. Step (iv): the Caratheodory set is replaced by its corresponding original points (dark blue). The remaining points in $P$ (bright blue) are deleted. Step (v): Previous steps are repeated until only $d+1=3$ points remains. This takes $O(\log n)$ iterations for $k=\Theta(d)$. This figure was taken from~\protect\cite{maalouf2019fast}.}
  \label{fig:booster}
\end{figure}

As explained in Section~\ref{sec:cara}, Algorithm~\ref{Alg:cara} takes as input a weighted set of $n$ points, and aims to compute a weighted subset of at most $d+1$ points which have the same weighted sum and whose weights sum to one. At each iteration, the algorithm sets at least one of the weights of the remaining points to zero. Hence, $O(n)$ such iterations are required to compute the required output, for a total running time of $O(n^2d^2)$.

However, it was suggested in~\cite{nasser2015coresets} that instead of taking as input the entire set of $n$ weighted points at once, it would be more efficient to run Algorithm~\ref{Alg:cara} in a streaming fashion as follows. Start with a set of $d+1$ weighted input points. Then, applying the following process $O(n)$ times: Add $1$ new weighted input point to the existing set and apply Caratheodorys theorem to reduce the $d+2$ points to $d+1$ points in $O(d^3)$ time. The running time of this algorithm is $O(nd^3)$ since it executed $n$ times the above procedure on only $d+2$ points.

Inspired by the streaming fashion of the previous algorithm, a more efficient algorithm was then proposed in~\cite{maalouf2019fast}. It suggests to do the following.

(i) Partition the input weighted set into $k \in O(n/d)$ subsets $P_1,\cdots,P_k$, each of size at most $O(d)$ (specifically $2d$).

(ii) Compute the weighted sum $\mu_i$ of each chunk $P_i$.

(iii) Apply the Caratheodory theorem only to the set $\mu = \br{\mu_1,\cdots,\mu_k}$ of $k$ weighted sums to obtain a weighted subset $\hat{\mu}$ of $|\hat{\mu}| = d+1$ elements from $\mu$.

(iv) Delete every chunk $P_j$ whose weighted mean $\mu_j$ was not chosen by the Caratheodory theorem (i.e., $\mu_j \not\in \hat{\mu}$).

(v) Continue recursively until only $d+1$ input points remain. This algorithm is illustrated in Fig.~\ref{fig:booster}.
The running time of this algorithm is $O(nd+d^4 \log{n})$; see Theorem~\ref{booster}.

The following theorem is a restatement of Theorem $3.1$ in~\cite{maalouf2019fast} for $k=2d$.
\begin{theorem}[Theorem $3.1$ in~\cite{maalouf2019fast}] \label{booster}
Let $(P,w)$ be a weighted set of $n$ points in $\REAL^d$ such that $w:P\to [0,1]$ and $\sum_{p\in P}w(p) = 1$. Then a weighted set $(S,u)$ that satisfies that $S\subseteq P$, $|S| \leq d+1$, $u:S\to [0,1]$, $\sum_{s\in S}u(s) = 1$ and $\sum_{s\in S}u(s)s = \sum_{p\in P}w(p)p$ can be computed in time $O\left(nd+d^4\log{n}\right)$.
\end{theorem}

\bibliography{references_accurate}

\appendix

\section{The Basics of Linear Algebra} \label{sec:LA}
In this section, we give a brief overview of tools and definitions from linear algebra which are used along the paper.

Recall that the $d\times d$ identity matrix is denoted by $I_d$.
An orthogonal matrix is a square matrix $M \in \REAL^{d\times d}$ whose columns and rows are orthogonal unit vectors, i.e. $M^TM = MM^T = I_d$.
The rank of a matrix $A \in \REAL^{m\times n}$ is the dimension of the its columns space, i.e., the vector space that is spanned by its columns, which is identical to the dimension of its rows space. This is also the maximal number of linearly independent columns of $A$. The matrix $A$ is said to have full rank if its rank equals the smaller value between $m$ and $n$.

The \emph{Gram–Schmidt} process or QR-decomposition~\cite{schmidt1908auflosung,golub1996matrix} is a method for orthonormalising a set of vectors in an inner product space (for example the Euclidean space $\REAL^d$ equipped with the standard inner product). The Gram–Schmidt process takes a finite, linearly independent set $S = \br{a_1,\cdots, a_k}$ of vectors for $k\leq d$ and generates an orthogonal set $S′ = \br{u_1,\cdots, u_k}$ that spans the same $k$-dimensional subspace of $\REAL^d$ as $S$. The application of the Gram–Schmidt process to the column vectors of a full column rank matrix $A \in \REAL^{m\times n}$ with $m \geq n$ yields the \emph{QR decomposition} $A = QR$~\cite{trefethen1997numerical} where $Q \in \REAL^{m\times m}$ is an orthogonal matrix and $R \in \REAL^{m\times n}$ is a right triangular matrix; see Fig.~\ref{fig:QRDecomp}.

A \emph{Singular Value Decomposition} (SVD) of a matrix~\cite{klema1980singular} $A\in \REAL^{m\times n}$ is a factorization $A = UDV^T$ such that $U \in \REAL^{m\times m}$ and $V\in \REAL^{n\times n}$ are orthogonal matrices and $D\in \REAL^{m\times n}$ is a diagonal matrix whose diagonal entries are called \emph{singular values} of $A$ and are non-negative and non-increasing; see Fig.~\ref{fig:SVD}.

A \emph{Thin Singular Value Decomposition} (thin SVD) of a matrix $A \in \REAL^{m\times n}$ of rank $r \leq n$ is a factorization $A = U_r D_r V_r^T$ such that $U_r \in \REAL^{m\times r}$, $V_r \in \REAL^{n\times r}$, $U_r^TU_r = I_r$, $V_r^TV_r = I_r$, and $D \in \REAL^{r\times r}$ is a diagonal matrix containing the singular values $d_1 \geq \cdots \geq d_r > 0$; see Fig.~\ref{fig:thinSVD}.

Every matrix $A \in \REAL^{m\times n}$ has a QR, SVD, and thin SVD decompositions.

Let $L \subseteq \REAL^d$ be a $j$-dimensional linear subspace and let $S \in \REAL^{d\times j}$ be a matrix whose columns are mutually orthogonal and span $L$. Let $S^\bot \in \REAL^{d\times (d-j)}$ be a matrix whose columns are mutually orthogonal and span the orthogonal complement $L^\bot$ of $L$. By the Pythagorean theorem, the squared Euclidean distance between a point $p \in \REAL^d$ and $L$ can be computed by taking the norm of its projection onto $A^\bot$, namely $\dist^2(p,L) = \norm{p}^2 - \norm{p^T S}^2 = \norm{p^T S^\bot}^2$. The sum of squared distances from the rows of a matrix $A \in \REAL^{n \times d}$ to $L$ is thus $\norm{A S^\bot}_F^2$.

A natural application for SVD is to compute the unit vector $x\in\REAL^n$ that minimizes $\norm{Ax}_2^2$~\cite{klema1980singular} $A\in \REAL^{m \times n}$ is a given matrix. The desired vector of coefficients $x$ is simply the vector in the matrix $V$ that corresponds to the smallest singular value in the SVD of the matrix $A$. Minimizing the over determined system $\norm{Ax-b}_2^2$ given an additional non-zero vector $b \in \REAL^m$ can be done as follows. Let $UDV^T$ be the SVD of $A$. We now have that $Ax = UDV^Tx = b$. Multiplying both sides by $U^T$ yields $DV^T = U^Tb$. Multiplying again by the pseudo inverse $D^\dag$ of $D$ yields $V^Tx = D^\dag (U^Tb)$. Hence, by multiplying both sides by $V$ we get that $x = V(D^\dag(U^Tb))$ is the desired vector of coefficients.

If the matrix $A$ contains at least $n$ independent vectors in its rows, then the rows of $A$ is a basis for $R^n$. This implies that every vector $v\in\REAL^n$ is a linear combination of only $n$ such vectors. Section~\ref{sec:LMS} suggests such a "coreset" with only positive coefficients in the linear combination. Computing these coefficients can be done by solving the linear system $Bx=v$ where the columns of $B$ are the vectors that span $v$.


\end{document}